\pdfoutput=1
\documentclass[11pt,a4paper]{article}
\usepackage[hyperref]{eacl2021}
\usepackage{times}
\usepackage{latexsym}

\usepackage{microtype}

\aclfinalcopy % Uncomment this line for the final submission
\usepackage{natbib}
\usepackage{array}
\usepackage{arydshln}
\usepackage{times}
\usepackage{comment}
\usepackage{url}
\usepackage{float}
\usepackage{amsfonts}
\usepackage{multirow}
\usepackage{xcolor}
\usepackage{csquotes}
\usepackage{graphicx}
\usepackage{amsmath}
\usepackage{diagbox}
\usepackage{booktabs,caption}
\usepackage[flushleft]{threeparttable}
\usepackage{times}
\usepackage{url}
\usepackage{latexsym}
\usepackage{arydshln}
\usepackage{latexsym}
\usepackage{comment}
\usepackage{float}
\usepackage{amsfonts}
\usepackage{multirow}
\usepackage{xcolor}
\usepackage{csquotes}
\usepackage{graphicx}
\usepackage{amsmath}
\usepackage{cleveref}

\crefformat{section}{\S#2#1#3} % see manual of cleveref, section 8.2.1
\crefformat{subsection}{\S#2#1#3}
\crefformat{subsubsection}{\S#2#1#3}

\title{Neural Supervised Domain Adaptation by Augmenting Pre-trained Models with Random Units}

\author{Sara Meftah$^{\ast}$, Nasredine Semmar$^{\ast}$, Youssef Tamaazousti$^{\ast}$, Hassane Essafi$^{\ast}$, Fatiha Sadat$^{+}$ \\
  $^{\ast}$CEA-List, Université Paris-Saclay, F-91120, Palaiseau, France\\ 
  $^{+}$UQ{\`A}M, Montr{\'e}al, Canada  \\
     {\tt \{firstname.lastname\}@cea.fr}, {\tt sadat.fatiha@uqam.ca } \\}         

\date{}

\begin{document}
\maketitle
\begin{abstract}
Neural Transfer Learning (TL) is becoming ubiquitous in Natural Language Processing (NLP), thanks to its high performance on many tasks, especially in low-resourced scenarios. Notably, TL is widely used for neural domain adaptation to transfer valuable knowledge from high-resource to low-resource domains. In the standard fine-tuning scheme of TL, a model is initially pre-trained on a source domain and subsequently fine-tuned on a target domain and, therefore, source and target domains are trained using the same architecture. In this paper, we show through interpretation methods that such scheme, despite its efficiency, is suffering from a main limitation. Indeed, although capable of adapting to new domains, pre-trained neurons struggle with learning certain patterns that are specific to the target domain. Moreover, we shed light on the hidden negative transfer occurring despite the high relatedness between source and target domains, which may mitigate the final gain brought by transfer learning. To address these problems, we propose to augment the pre-trained model with normalised, weighted and randomly initialised units that foster a better adaptation while maintaining the valuable source knowledge. We show that our approach exhibits significant improvements to the standard fine-tuning scheme for neural domain adaptation from the news domain to the social media domain on four NLP tasks: part-of-speech tagging, chunking, named entity recognition and morphosyntactic tagging.\footnote{Under review}
\end{abstract}

%=========================
% Introduction
%=========================
\section{Introduction}
\label{sec:introduction}

NLP aims to produce resources and tools to understand texts coming from standard languages and their linguistic varieties, such as dialects or user-generated-content in social media platforms. This diversity is a challenge for developing high-level tools that are capable of understanding and generating all forms of human languages. Furthermore, in spite of the tremendous empirical results achieved by NLP models based on Neural Networks (NNs), these models are in most cases based on a supervised learning paradigm, \textit{i.e.} trained from scratch on large amounts of labelled examples. Nevertheless, such training scheme is not fully optimal. Indeed, NLP neural models with high performance often require huge volumes of manually annotated data to produce powerful results and prevent over-fitting. However, manual data annotation is time-consuming. Besides, language changes over years \citep{eisenstein2019measuring}. Thus, most languages varieties are under-resourced \citep{baumann2014using,duong2017natural}. 

Particularly, in spite of the valuable advantage of social media's content analysis for a variety of applications (\textit{e.g.} advertisement, health, or security), this large domain is still poor in terms of annotated data. Furthermore, it has been shown that models intended for news fail to work efficiently on Tweets \citep{owoputi2013improved}. This is mainly due to the conversational nature of the text, the lack of conventional orthography, the noise, linguistic errors, spelling inconsistencies, informal abbreviations and the idiosyncratic style of these texts \citep{horsmann2018robust}.

One of the best approaches to address this issue is Transfer Learning (TL); an approach that allows handling the problem of the lack of annotated data, whereby relevant knowledge previously learned in a source problem is leveraged to help in solving a new target problem \citep{pan2010survey}. In the context of artificial NNs, TL relies on a model learned on a \textit{source-task} with sufficient data, further adapted to the \textit{target-task} of interest. TL has been shown to be powerful for NLP and outperforms the standard supervised learning from scratch paradigm, because it takes benefit from the pre-learned knowledge. Particularly, the standard fine-tuning (SFT) scheme of sequential transfer learning has been shown to be efficient for supervised domain adaptation from the source news domain to the target social media domain \citep{gui2017part,meftah2018using,meftah2018neural,marz2019domain,zhao2017deep,lin2018neural}.

In this work we first propose a series of analysis to spot the limits of the standard fine-tuning adaptation scheme of sequential transfer learning. We start by taking a step towards identifying and analysing the \textit{hidden} negative transfer when transferring from the news domain to the social media domain. Negative transfer \citep{rosenstein2005transfer,wang2019characterizing} occurs when the knowledge learnt in the source domain hampers the learning of new knowledge from the target domain. Particularly, when the source and target domains are dissimilar, transfer learning may fail and hurt the performance, leading to a worse performance compared to the standard supervised training from scratch. In this work, we rather perceive the gain brought by the standard fine-tuning scheme compared to random initialisation\footnote{Random initialisation means training from scratch on target data (in-domain data).} as a combination of a positive transfer and a hidden negative transfer. We define positive transfer as the percentage of predictions that were wrongly predicted by random initialisation, but using transfer learning changed to the correct ones. The negative transfer represents the percentage of predictions that were tagged correctly by random initialisation, but using transfer learning gives incorrect predictions. Hence, the final gain brought by transfer learning would be the difference between positive and negative transfer. We show that despite the final positive gain brought by transfer learning from the high-resource news domain to the low-resource social media domain, the \textit{hidden negative transfer} may mitigate the final gain.

Then we perform an interpretive analysis of individual pre-trained neurons behaviours in different settings. We find that some of pretrained neurons are \textit{biased} by what they have learnt in the source-dataset. For instance, we observe a unit\footnote{We use \enquote{unit} and \enquote{neuron} interchangeably.} firing on proper nouns (\textit{e.g.}\enquote{George} and \enquote{Washington}) before fine-tuning and on words with capitalised first-letter whether the word is a proper noun or not (\textit{e.g.} \enquote{Man} and \enquote{Father}) during fine-tuning. Indeed, in news, only proper nouns start with an upper-case letter. Thus the pre-trained units fail to discard this pattern which is not always respected in user-generated-content in social media. As a consequence of this phenomenon, specific patterns to the target-dataset (\textit{e.g.} ``wanna'' or ``gonna'') are difficult to learn by pre-trained units. This phenomenon is non-desirable, since such specific units are essential, especially for target-specific classes \citep{zhou2018revisiting,lakretz2019emergence}.

Stemming from our analysis, we propose a new method to overcome the above-mentioned drawbacks of the standard fine-tuning scheme of transfer learning. Precisely, we propose a hybrid method that takes benefit from both worlds, random initialisation and transfer learning, without their drawbacks. It consists in augmenting the source-network (set of pre-trained units) with randomly initialised units (that are by design non-biased) and jointly learn them.  We call our method \textbf{PretRand} (\textbf{Pret}rained and \textbf{Rand}om units). PretRand consists of three main ideas:

\begin{enumerate}
    \item Augmenting the source-network (set of pre-trained units) with a random branch composed of randomly initialised units, and jointly learn them.
    
    \item Normalising the outputs of both branches to balance their different behaviours and thus forcing the network to consider both.
    
    \item Applying learnable attention weights on both branches predictors to let the network learn which of random or pre-trained one is better for every class. 
    
\end{enumerate}

Our experiments on 4 NLP tasks: Part-of-Speech tagging (POS), Chunking (CK), Named Entity Recognition (NER) and Morphosyntactic Tagging (MST) show that PretRand enhances considerably the performance compared to the standard fine-tuning adaptation scheme.\footnote{This paper is an extension of our previous work \citep{meftah2019joint}.}

The remainder of this paper is organised as follows. Section \ref{sec:background} presents the background related to our work: transfer learning and interpretation methods for NLP. Section~\ref{sec:base_model} presents the base neural architecture used for sequence labelling in NLP. Section~\ref{sec:sft_scheme} describes our proposed methods to analyse the standard fine-tuning scheme of sequential transfer learning. Section~\ref{sec:proposed_method} describes our proposed approach PretRand. Section~\ref{sec:experiments} reports the datasets and the experimental setup. Section~\ref{sec:results} reports the experimental results of our proposed methods and is divided into two sub-sections: Sub-section~\ref{sec:analysis_sft_results} reports the empirical analysis of the standard fine-tuning scheme, highlighting its drawbacks. Sub-section~\ref{sec:pretrand_results} presents the experimental results of our proposed approach PretRand, showing the effectiveness of PretRand on different tasks and datasets and the impact of incorporating contextualised representations. Finally, section~\ref{sec:conclusion} wraps up by discussing our findings and future research directions.

%=========================
% Related Work
%=========================
\section{Background}
\label{sec:background} 

Since our work involves two research topics: Sequential Transfer Learning (STL) and Interpretation methods, we discuss in the following sub-sections the state-of-the-art of each topic with a positioning of our contributions regarding each one.

%=========================
% Related Work: Sequential Transfer Learning
%=========================
\subsection{Sequential Transfer Learning}
\label{sec:related_work_Sequential_Transfer_Learning}

In STL, training is performed in two stages, sequentially: \textbf{pretraining} on the source task, followed by an \textbf{adaptation} on the downstream target tasks \citep{ruder2019neural}. The purpose behind using STL techniques for NLP can be divided into two main research areas, \textbf{universal representations} and \textbf{domain adaptation}. 

Universal representations aim to build neural features (\textit{e.g.} words embeddings and sentence embeddings) that are transferable and beneficial to a wide range of downstream NLP tasks and domains. Indeed, the probabilistic language model proposed by \citet{bengio2003neural} was the genesis of what we call words embedding in NLP, while Word2Vec \citep{mikolov2013efficient} was its outbreak and a starting point for a surge of works on learning words embeddings: \textit{e.g.} FastText \citep{bojanowski2017enriching} enriches Word2Vec with subword information. Recently, universal representations re-emerged with contextualised representations, handling a major drawback of traditional words embedding. Indeed, these last learn a single context-independent representation for each word thus ignoring words polysemy. Therefore, contextualised words representations aim to learn context-dependent word embeddings, \textit{i.e.} considering the entire sequence as input to produce each word's embedding. 

While universal representations seek to be propitious for any downstream task, domain adaptation is designed for particular target tasks. Domain adaptation consists in adapting NLP models designed for specific high-resourced source setting(s) (language, language variety, domain, task, etc) to work in a target low-resourced setting(s). It includes two categories. First, unsupervised domain adaptation assumes that labelled examples in the source domain are sufficiently available, but for the target domain, only unlabelled examples are available. Second, in supervised domain adaptation setting, a small number of labelled target examples are assumed to be available. \\

%=========================
% Related Work: Pretraining
%=========================
\noindent
\textbf{Pretraining}

\noindent
In the pretraining stage of STL, a crucial key for the success of transfer is the ruling about the pre-trained task and domain. For \textit{universal representations}, the pre-trained task is expected to encode useful features for a wide number of target tasks and domains. In comparison, for \textit{domain adaptation}, the pre-trained task is expected to be most suitable for the target task in mind. We classify pretraining methods into four main categories: unsupervised, supervised, multi-task and adversarial pretraining: 

\begin{itemize}
    \item \textit{Unsupervised pretraining} uses raw unlabelled data for pretraining. Particularly, it has been successfully used in a wide range of seminal works to learn universal representations. Language modelling task has been particularly used thanks to its ability to capture general-purpose features of language.\footnote{Note that language modelling is also considered as a self-supervised task since, in fact, labels are automatically generated from raw data.} For instance, TagLM~\citep{peters2017semi} is a pretrained model based on a bidirectional language model (biLM), also used to generate ELMo (Embeddings from Language Models) representations~\citep{peters2018deep}. With the recent emergence of the \enquote{Transformers} architectures~\citep{vaswani2017attention}, many works propose pretrained models based on these architectures~\citep{devlin2019bert,yang2019xlnet,2019t5}. Unsupervised pretraining has also been used to improve sequence to sequence learning. We can cite the work of~\citet{ramachandran2017unsupervised} who proposed to improve the performance of an encoder-decoder neural machine translation model by initialising both encoder and decoder parameters with pretrained weights from two language models.

    \item \textit{Supervised pretraining} has been particularly used for cross-lingual transfer (\textit{e.g.} machine translation~\citep{zoph2016multi}), cross-task transfer from POS tagging to words segmentation task \citep{yang2017neural} and cross-domain transfer for biomedical texts for question answering by \citet{wiese2017neural} and for NER by \citet{giorgi2018transfer}. Cross-domain transfer has also been used to transfer from news to social media texts for POS tagging \citep{meftah2017supervised,marz2019domain} and sentiment analysis \citep{zhao2017deep}. Supervised pretraining has been also used effectively for universal representations learning, \textit{e.g.} neural machine translation \citep{mccann2017learned}, language inference \citep{conneau2017supervised} and discourse relations \citep{nie2017dissent}. 

    \item \textit{Multi-task pretraining} has been successfully applied to learn general universal sentence representations by a simultaneous pretraining on a set of supervised and unsupervised tasks~\citep{subramanian2018learning,cer2018universal}. \citet{subramanian2018learning}, for instance, proposed to learn universal sentences representations by a joint pretraining on skip-thoughts, machine translation, constituency parsing, and natural language inference. For domain adaptation, we have performed in~\citep{meftah2020multi} a multi-task pretraining for supervised domain adaptation from the news domain to the social media domain.

    \item \textit{Adversarial pretraining} is particularly used for domain adaptation when some annotated examples from the target domain are available. Adversarial training \citep{ganin2016domain} is used as a pretraining step followed by an adaptation step on the target dataset. Adversarial pretraining demonstrated its effectiveness in several NLP tasks, \textit{e.g.} cross-lingual sentiment analysis~\citep{chen2018adversarial}. Also, it has been used to learn cross-lingual words embeddings~\citep{lample2018word}.

\end{itemize}

%=========================
% Related Work: Adaptation
%=========================
\noindent
\textbf{Adaptation}

\noindent
During the adaptation stage of STL, one or more layers from the pretrained model are transferred to the downstream task, and one or more randomly initialised layers are added on top of pretrained ones. Three main adaptation schemes are used in sequential transfer learning: \textit{Feature Extraction}, \textit{Fine-Tuning} and the recent \textit{Residual Adapters}. 

In a \textit{Feature Extraction} scheme, the pretrained layers' weights are frozen during adaptation, while in \textit{Fine-Tuning} scheme weights are tuned. Accordingly, the former is computationally inexpensive while the last allows better adaptation to target domains peculiarities. In general, fine-tuning pretrained models begets better results, except in cases wherein the target domain's annotations are sparse or noisy \citep{dhingra2017comparative,mou2016transferable}. \citet{peters2019tune} found that for contextualised representations, both adaptation schemes are competitive, but the appropriate adaptation scheme to pick depends on the similarity between the source and target problems. Recently, \textit{Residual Adapters} were proposed by \citet{houlsby2019parameter} to adapt pretrained models based on Transformers architecture, aiming to keep \textit{Fine-Tuning} scheme's advantages while reducing the number of parameters to update during the adaptation stage. This is achieved by adding adapters (intermediate layers with a small number of parameters) on top of each pretrained layer. Thus, pretrained layers are frozen, and only adapters are updated during training.
Therefore,  \textit{Residual Adapters} performance is near to \textit{Fine-tuning} while being computationally cheaper \citep{pfeiffer2020adapterhub,pfeiffer2020adapterfusion,pfeiffer2020mad}.

\noindent
\textbf{Our work}

\noindent
Our work falls under \textit{supervised domain adaptation} research area. Specifically, cross-domain adaptation from the news domain to the social media domain. The fine-tuning adaptation scheme has been successfully applied on domain adaptation from the news domain to the social media domain (\textit{e.g.} adversarial pretraining \citep{gui2017part} and supervised pretraining \citep{meftah2018neural}). In this research, we highlight the aforementioned drawbacks (biased pre-trained units and the hidden negative transfer) of the standard fine-tuning adaptation scheme. Then, we propose a new adaptation scheme (PretRand) to handle these problems. Furthermore, while ELMo contextualised words representations efficiency has been proven for different tasks and datasets \citep{peters2019tune,fecht2019sequential,schumacher2019learning}, here we investigate their impact when used, simultaneously, with a sequential transfer learning scheme for supervised domain adaptation.

%=========================
% Related Work: Visualization methods for NLP
%=========================
\subsection{Interpretation methods for NLP}
\label{sec:related_work_visualization_methods_for_NLP} 

Recently, a rising interest is devoted to peek inside black-box neural NLP models to interpret their internal representations and their functioning. A variety of methods were proposed in the literature, here we only discuss those that are most related to our research.\\

\noindent
\textbf{Probing tasks} is a common approach for NLP models analysis used to investigate which linguistic properties are encoded in the latent representations of the neural model \citep{shi2016does}. Concretely, given a neural model $\mathbf{M}$ trained on a particular NLP task, whether it is unsupervised (\textit{e.g.} language modelling (LM)) or supervised (\textit{e.g.} Neural Machine Translation (NMT)), a shallow classifier is trained on top of the frozen $\mathbf{M}$ on a corpus annotated with the linguistic properties of interest. The aim is to examine whether $\mathbf{M}$'s hidden representations encode the property of interest. For instance, \citet{shi2016does} found that different levels of syntactic information are learned by NMT encoder's layers. \citet{adi2016fine} investigated what information (between sentence length, words order and word-content) is captured by different sentence embedding learning methods. \citet{conneau2018you} proposed 10 probing tasks annotated with fine-grained linguistic properties and compared different approaches for sentence embeddings. \citet{zhu2018exploring} inspected which semantic properties (\textit{e.g.} negation, synonymy, etc.) are encoded by different sentence embeddings approaches. Furthermore, the emergence of contextualised words representations have triggered a surge of works on probing what these representations are learning \citep{liu2019linguistic,clark2019does}. This approach, however, suffers from two main flaws. First, probing tasks examine properties captured by the model at a coarse-grained level, \textit{i.e.} layers representations, and thereby, will not identify features captured by individual neurons. Second, probing tasks will not identify linguistic properties that do not appear in the annotated probing datasets \citep{zhou2018interpreting}.\\

\noindent
\textbf{Individual units stimulus}: Inspired by works on receptive fields of biological neurons \citep{hubel1965receptive}, much work has been devoted for interpreting and
visualising individual hidden units stimulus-features in neural networks. Initially, in computer vision \citep{coates2011selecting,girshick2014rich,zhou2014object}, and more recently in NLP, wherein units activations are visualised in heatmaps. For instance, \citet{karpathy2016visualizing} visualised character-level Long Short-Term Memory (LSTM) cells learned in language modelling and found multiple interpretable units that track long-distance dependencies, such as line lengths and quotes; \citet{radford2017learning} visualised a unit which performs sentiment analysis in a language model based on Recurrent Neural Networks (RNNs); \citet{bau2018identifying} visualised neurons specialised on tense, gender, number, etc. in NMT models; and \citet{kadar2017representation} proposed \textit{top-k-contexts} approach to identify sentences, an thus linguistic patterns, sparking the highest activation values of each unit in an RNNs-based model. \\

\noindent
\textbf{Neural representations correlation analysis}: Cross-network and cross-layers correlation is a significant approach to gain insights on how internal representations may vary across networks, network-depth and training time. Suitable approaches are based on Correlation Canonical Analysis (CCA) \citep{hotelling1992relations,uurtio2018tutorial}, such as Singular Vector Canonical Correlation Analysis \citep{raghu2017svcca} and Projected Weighted Canonical Correlation Analysis \citep{morcos2018insights}, that were successfully used in NLP neural models analysis. For instance, it was used by \citet{bau2018identifying} to calculate cross-networks correlation for ranking important neurons in NMT and LM. \citet{saphra2019understanding} applied it to probe the evolution of syntactic, semantic, and topic representations cross-time and cross-layers. \citet{raghu2019transfusion} compared the internal representations of models trained from scratch \textit{vs} models initialised with pre-trained weights. CCA based methods aim to calculate similarity between neural representations at the coarse-grained level. In contrast, correlation analysis at the fine-grained level, \textit{i.e.} between individual neurons, has also been explored in the literature. Initially, \citet{li2015convergent} used Pearson's correlation to examine to which extent each individual unit is correlated to another unit, either within the same network or between different networks. The same correlation metric was used by \citet{bau2018identifying} to determine important neurons in NMT and LM tasks. \\

\noindent
\textbf{Our Work:}

\noindent
In this work, we propose two approaches (\cref{sec:proposed_visualization}) to highlight the bias effect in the standard fine-tuning scheme of transfer learning in NLP, the first method is based on \textit{individual units stimulus} and the second on \textit{neural representations correlation analysis}. To the best of our knowledge, we are the first to harness these interpretation methods to analyse individual units behaviour in a transfer learning scheme. Furthermore, the most analysed tasks in the literature are Natural Language Inference, NMT and LM \citep{belinkov2019analysis}, here we target under-explored tasks in visualisation works such as POS, MST, CK and NER.

%=========================
% Base Model
%=========================
\section{Base Neural Sequence Labelling Model}
\label{sec:base_model} 

%=========================
% Base Sequence Labelling model architecture
%=========================
\begin{figure*}
	\centering
		\includegraphics[scale=0.073]{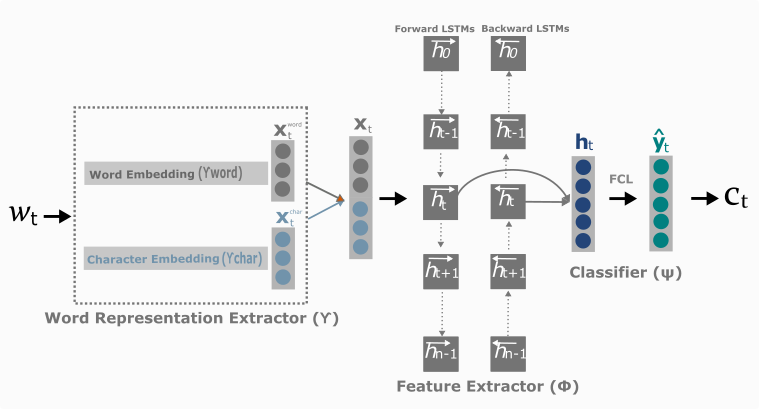}
	\caption{Illustrative scheme of the base neural model for sequence labelling tasks.}
	\label{fig:base_architecture}
\end{figure*}
%=========================

Given an input sentence $S$ of $n$ successive tokens $S=[w_{1},\ldots,w_{n}]$, the goal of sequence labelling is to predict the label $c_t$~$\in$~$\mathcal{C}$ of every $w_{t}$, with $\mathcal{C}$ being the tag-set. We use a commonly used end-to-end neural sequence labelling model \citep{ma2016end,plank-etal-2016-multilingual,yang2018design}, which is composed of three components (illustrated in Figure \ref{fig:base_architecture}). First, the \textbf{Word Representation Extractor}~($\mathbf{WRE}$), denoted $\Upsilon$, computes a vector representation $\mathbf{x}_t$ for each token \(w_t\). Second, this representation is fed into a \textbf{Feature Extractor}~($\mathbf{FE}$) based on a bidirectional Long Short-Term Memory (biLSTM) network \citep{graves2013hybrid}, denoted $\Phi$. It produces a hidden representation, $\mathbf{h}_t$, that is fed into a \textbf{Classifier}~($\mathbf{Cl}$): a fully-connected layer (FCL), denoted $\Psi$. Formally, given $w_t$, the logits are obtained using the following equation: $\hat{\mathbf{y}}_{t} = (\Psi \circ \Phi \circ \Upsilon)(w_t)$.\footnote{For simplicity, we define $\hat{\mathbf{y}}_{t}$ only as a function of $w_t$. In reality, the prediction $\hat{\mathbf{y}}_{t}$ for the word $w_t$ is also a function of the remaining words in the sentence and the model's parameters, in addition to $w_t$.} 

In the standard supervised training scheme, the three modules are jointly trained from scratch by minimising the Softmax Cross-Entropy (SCE) loss using the Stochastic Gradient Descent (SGD) algorithm. 

Let us consider a training set of $M$ annotated sentences, where each sentence $i$ is composed of $m_i$ tokens. Given a training word $(w_{i,t},y_{i,t})$ from the training sentence $i$, where $y_{i,t}$ is the gold standard label for the word $w_{i,t}$, the cross-entropy loss for this example is calculated as follows:

\begin{equation}
    \mathcal{L}^{(i,t)}~=~-~y_{i,t}~\times~log(\hat{y}_{i,t})~.
\end{equation}

\noindent
Thus, during the training of the sequence labelling model on $M$ annotated sentences, the model's loss is defined as follows:

\begin{equation}
    \mathcal{L}~=~\sum_{i=1}^{M}~\sum_{t=1}^{m_i}~\mathcal{L}^{(i,t)}~.
\end{equation}

%=========================
% Base Model
%=========================
\section{Analysis of the Standard Fine-Tuning Scheme}
\label{sec:sft_scheme}

The standard fine-tuning scheme consists in transferring a part of the learned weights from a source model to initialise the target model, which is further fine-tuned on the target task with a small number of training examples from the target domain. Given a source neural network $\mathcal{M}_{s}$ with a set of parameters $\theta_{s}$ split into two sets: $\theta_{s}=(\theta_{s}^1,\theta_{s}^2)$ and a target network $\mathcal{M}_{t}$ with a set of parameters $\theta_{t}$ split into two sets: $\theta_{t}=(\theta_{t}^1,\theta_{t}^2)$, the standard fine-tuning scheme of transfer learning includes three simple yet effective steps:

\begin{enumerate}
    \item We train the source model on annotated data from the source domain on a source dataset.
    \item We transfer the first set of parameters from the source network $\mathcal{M}_{s}$ to the target network $\mathcal{M}_{t}$:  $\theta_{t}^1 = \theta_{s}^1$, whereas the second set $\theta_{t}^2$ of parameters is randomly initialised. 
    \item Then, the target model is further fine-tuned on the small target data-set.
\end{enumerate}

Source and target datasets may have different tag-sets, even within the same NLP task. Hence, transferring the parameters of the classifier ($\Psi$) may not be feasible in all cases. Therefore, in our experiments, WRE's layers ($\Upsilon$) and FE's layers ($\Phi$) are initialised with the source model's weights and $\Psi$ is randomly initialised. Then, the three modules are further jointly trained on the \textit{target-dataset} by minimising a SCE loss using the SGD algorithm.

%=========================
% Proposed Method
%=========================
\subsection{The Hidden Negative Transfer}
\label{sec:proposed_method_negative_transfer}

It has been shown in many works in the literature \citep{rosenstein2005transfer,ge2014handling,ruder2019neural,gui2018negative,cao2018partial,chen2019catastrophic,wang2019characterizing,o2019learning} that, when the source and target domains are less related (\textit{e.g.} languages from different families), sequential transfer learning may lead to a negative effect on the performance, instead of improving it. This phenomenon is referred to as \textit{negative transfer}. Precisely, negative transfer is considered when transfer learning is harmful to the target task/dataset, \textit{i.e.} the performance when using transfer learning algorithm is lower than that with a solely supervised training on in-target data \citep{torrey2010transfer}. 

In NLP, negative transfer phenomenon has only seldom been studied. We can cite the recent work of \citet{kocmi2020exploring} who evaluated the negative transfer in transfer learning in neural machine translation when the transfer is performed between different language-pairs. They found that: 1)~The distributions mismatch between source and target language-pairs does not beget a negative transfer. 2)~The transfer may have a negative impact when the source language-pair is less-resourced compared to the target one, in terms of annotated examples.

Our experiments in \citep{meftah2018neural,meftah2018using} have shown that transfer learning techniques from the news domain to the social media domain using the standard fine-tuning scheme boosts the tagging performance. Hence, following the above definition, transfer learning from news to social media does not beget a negative transfer. Contrariwise, in this work, we instead consider the \textit{hidden negative transfer}, \textit{i.e.} the percentage of predictions that were correctly tagged by random initialisation, but using transfer learning gives wrong predictions.

Let us consider the gain $\mathcal{G}_{i}$ brought by the standard fine-tuning scheme (SFT) of transfer learning compared to the random initialisation for a dataset ${i}$. $\mathcal{G}_{i}$ is defined as the difference between positive transfer $\mathcal{PT}_{i}$ and negative transfer $\mathcal{NT}_{i}$:

\begin{equation}
\mathcal{G}_{i}= \mathcal{PT}_{i} - \mathcal{NT}_{i},
\end{equation}

\noindent
where positive transfer $\mathcal{PT}_{i}$ represents the percentage of tokens that were wrongly predicted by random initialisation, but the SFT changed to the correct ones. \textit{Negative transfer} $\mathcal{NT}_{i}$ represents the percentage of words that were tagged correctly by random initialisation, but using SFT gives wrong predictions. $\mathcal{PT}_{i}$ and $\mathcal{NT}_{i}$ are defined as follows:

\begin{equation}
\mathcal{PT}_{i}= \frac{N^{corrected}_{i}}{N_{i}},
\end{equation}

\begin{equation}
\mathcal{NT}_{i}= \frac{N^{falsified}_{i}}{N_{i}},
\end{equation}

\noindent
where $N_{i}$ is the total number of tokens in the validation-set, $N^{corrected}_{i}$ is the number of tokens from the validation-set that were wrongly tagged by the model trained from scratch but are correctly predicted by the SFT scheme, and $N^{falsified}_{i}$ is the number of tokens from the validation-set that were correctly tagged by the model trained from scratch but are wrongly predicted by the SFT scheme.

%=========================
% Proposed Visualisation
%=========================
\subsection{Interpretation of Pretrained Neurons}
\label{sec:proposed_visualization} 

Here, we propose to perform a set of analysis techniques to gain some insights into how the inner pretrained representations are updated during fine-tuning on social media datasets when using the standard fine-tuning scheme of transfer learning. For this, we propose to analyse the feature extractor's ($\Phi$) activations. Precisely, we attempt to visualise \textit{biased neurons}, \textit{i.e.} pre-trained neurons that do not change that much from their initial state.

Let us consider a validation-set of $N$ words, the feature extractor $\Phi$ generates a matrix ${\mathbf{h}}~\in~{\mathbf{M}_{N,H}(\mathbb{R})}$ of activations over all the words of the validation-set, where $\mathbf{\mathbf{M}_{f,g}(\mathbb{R})}$ is the space of $f~\times~g$ matrices over $\mathbb{R}$ and $H$ is the size of the hidden representation (number of neurons). Each element $h_{i,j}$ from the matrix represents the activation of the neuron $j$ on the word $w_i$. 

Given two models, the first before fine-tuning and the second after fine-tuning, we obtain two matrices $\mathbf{h}^{before}~\in~{\mathbf{M}_{N,H}(\mathbb{R})}$ and $\mathbf{h}^{after}~\in~{\mathbf{M}_{N,H}(\mathbb{R})}$, which give the activations of $\Phi$ over all validation-set's words before and after fine-tuning, respectively. 

We aim to visualise and quantify the change of the representations generated by the model from the initial state, $\mathbf{h}^{before}$ (before fine-tuning), to the final state, $\mathbf{h}^{after}$ (after fine-tuning). For this purpose, we perform two experiments: 

\begin{enumerate}
    \item Quantifying the change of pretrained individual neurons (\cref{sec:proposed_visualization_correlation});
    \item Visualising the evolution of pretrained neurons stimulus during fine-tuning (\cref{sec:proposed_visualization_units_topk}).
\end{enumerate}

%=========================
% Sec: Correlation Analysis
%=========================
\subsubsection{Quantifying the change of individual pretrained neurons}
\label{sec:proposed_visualization_correlation} 

In order to quantify the change of the knowledge encoded in pretrained neurons after fine-tuning, we propose to calculate the similarity (correlation) between neurons activations before and after fine-tuning, when using the SFT adaptation scheme. Precisely, we calculate the correlation coefficient between each neuron's activations on the target-domain validation-set before starting fine-tuning and at the end of fine-tuning. 

%-------------------------
% Scheme activations method
%-------------------------
\begin{figure*}
	\centering
		\includegraphics[scale=0.075]{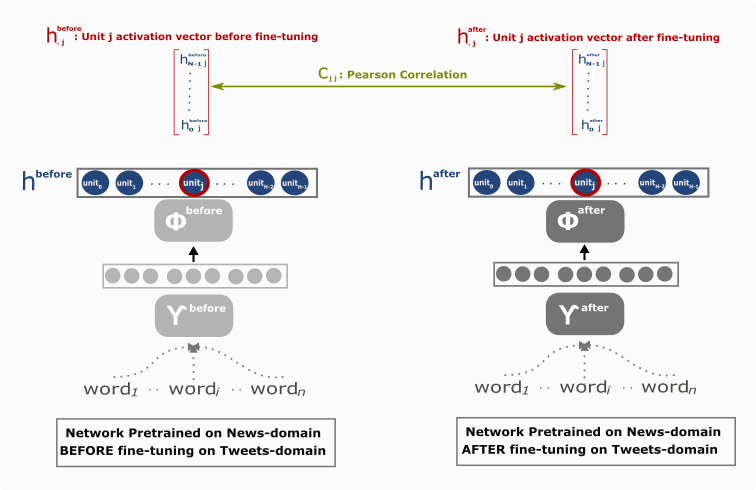}
	\caption{Illustrative scheme of the computation of the \textit{charge} of unit $j$, \textit{i.e.} the Pearson correlation between unit $j$ activations vector after fine-tuning to its activations vector before fine-tuning.}
	\label{fig:pearson_correlation_method}
\end{figure*}
%---------------------------
%-------------------------
% Scheme top k
%-------------------------

\begin{figure*}
	\centering
		\includegraphics[scale=0.07]{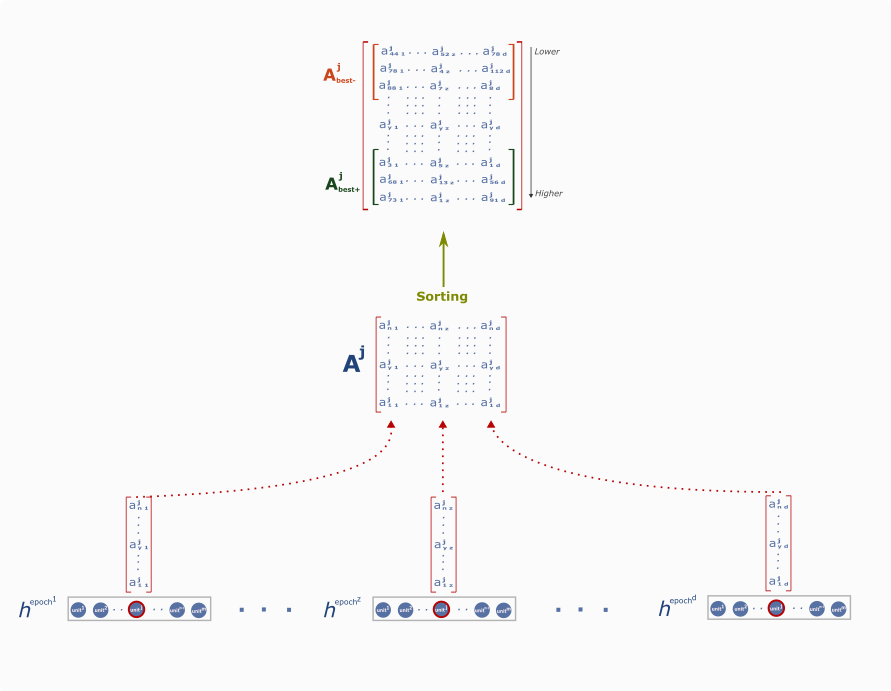}
	\caption{Illustrative scheme of the calculus of top-k-words activating unit $j$, positively ($\mathbf{A}^{(j)}_{best+}$) and negatively ($\mathbf{A}^{(j)}_{best-}$) during fine-tuning epochs. $\mathbf{h}^{epoch^z}$ states for $\Phi$'s outputs at epoch number $z$.}
	\label{fig:top_k_words_method}

\end{figure*}

%---------------------------

Following the above formulation and as illustrated in Figure \ref{fig:pearson_correlation_method}, from $\mathbf{h}^{before}$ and $\mathbf{h}^{after}$ matrices, we extract two vectors $\mathbf{h}^{before}_{.j}~\in$~$\mathbb{R}^{N}$ and $\mathbf{h}^{after}_{.j}\in$~$\mathbb{R}^{N}$, representing respectively the activations of a unit $j$ over all validation-set’s words \textit{before} and \textit{after} fine-tuning. Next, we generate an asymmetric correlation matrix $\mathbf{C}\in$~${\mathbf{M}_{H,H}(\mathbb{R})}$, where each element $\mathbf{c}_{jt}$ in the matrix represents the Pearson's correlation between the activation vector of unit $j$ after fine-tuning ($\mathbf{h}^{after}_{.j}$) and the activation vector of unit $t$  before fine-tuning ($\mathbf{h}^{before}_{.t}$), computed as follows:

\begin{equation}
\mathbf{c}_{jt}=  \frac{\mathbb{E}[(\mathbf{h}^{after}_{.j}-\mu^{after}_{j})(\mathbf{h}^{before}_{.t}-\mu^{before}_{t})]}{\sigma^{after}_{j}\sigma^{before}_{t}}~.
\end{equation}

\noindent
Here $\mu^{before}_{j}$ and $\sigma^{before}_{j}$ represent, respectively, the mean and the standard deviation of unit $j$ activations over the validation set. Clearly, we are interested by the matrix diagonal, where $c_{jj}$ represents the \textit{charge} of each unit $j$ from $\Phi$, \textit{i.e.} the correlation between each unit's activations after fine-tuning to its activations before fine-tuning.

%=========================
% Sec: Individual units visualisation during adaptation
%=========================
\subsubsection{Visualising the Evolution of Pretrained Neurons Stimulus during Fine-tuning}
\label{sec:proposed_visualization_units_topk}

Here, we perform units visualisation at the individual-level to gain insights on how the patterns encoded by individual units progress during fine-tuning when using the SFT scheme. To do this, we generate top-k activated words by each unit; \textit{i.e.} words in the validation-set that fire the most the said unit, positively and negatively (since LSTMs generate positive and negative activations). In \citep{kadar2017representation}, top-k activated contexts from the model were plotted at the end of training (the best model), which shows on what each unit is specialised, but it does not give insights about how the said unit is evolving and changing during training. Thus, taking into account only the final state of training does not reveal the whole picture. Here, we instead propose to generate and plot top-k words activating each unit throughout the adaptation stage. We follow two main steps (as illustrated in Figure \ref{fig:top_k_words_method}):

\begin{enumerate}
    \item We represent each unit $j$ from $\Phi$ with a random matrix $\mathbf{A}^{(j)}\in$~${\mathbf{M}_{N,D}(\mathbb{R})}$ of the said unit's activations on all the validation-set at different training epochs, where $D$ is the number of epochs and $N$ is the number of words in the validation-set. Thus, each element $a^{(j)}_{y,z}$ represents the activation of the unit $j$ on the word $w_y$ at the epoch $z$.
    
    \item  We carry out a sorting of each column of the matrix (each column represents an epoch) and pick the higher \textit{k} words (for \textit{top-k} words firing the unit positively) and the lowest \textit{k} words (for \textit{top-k} words firing the unit negatively), leading to two matrices, $\mathbf{A}^{(j)}_{best+}\in$~$\mathbf{\mathbf{M}_{D,k}(\mathbb{R})}$ and $\mathbf{A}^{(j)}_{best-}\in$~$\mathbf{\mathbf{M}_{D,k}(\mathbb{R})}$, the first for \textit{top-k} words activating positively the unit $j$ at each training epoch, and the last for \textit{top-k} words activating negatively the unit $j$ at each training epoch. 
    
\end{enumerate}

\newpage
%=========================
% Proposed Method
%=========================
\section{Joint Learning of Pretrained and Random Units: PretRand}
\label{sec:proposed_method}

\begin{figure*}
\centering
\centerline{\includegraphics[scale=0.07]{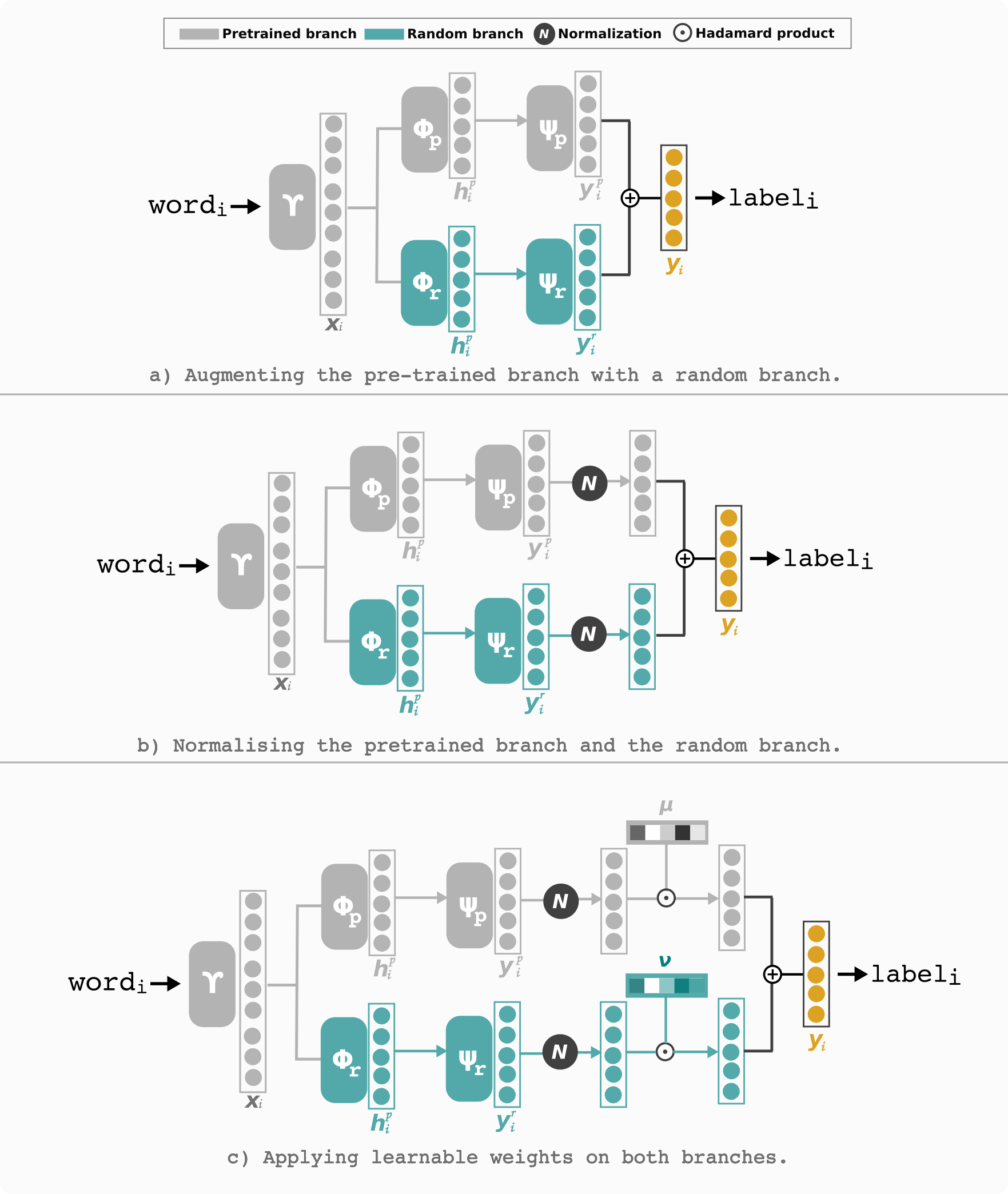}}
\caption{\textbf{Illustrative scheme of the three ideas composing our proposed adaptation method, PretRand.} a)~We augment the pre-trained branch (grey branch) with a randomly initialised one (green branch) and jointly adapt them with pre-trained ones (grey branch). An element-wise sum is further applied to merge the two branches. b)~Before merging, we balance the different behaviours of pre-trained and random units, using an independent normalisation ($\mathbf{N}$). c)~Finally we let the network learn which of pre-trained or random neurons are more suited for every class, by performing an element-wise product of the FC layers with learnable weighting vectors ($\mathbf{u}$ and $\mathbf{v}$ initialised with 1-values). }
\label{fig:pretrand}

\end{figure*}

We found from our analysis (in section \ref{sec:analysis_sft_results}) on pre-trained neurons behaviours, that the standard fine-tuning scheme suffers from a main limitation. Indeed, some pre-trained neurons still biased by what they have learned from the source domain despite the fine-tuning on target domain. We thus propose a new adaptation scheme, \textbf{PretRand}, to take benefit from both worlds, the pre-learned knowledge in the pretrained neurons and the target-specific features easily learnt by random neurons. PretRand, illustrated in Figure \ref{fig:pretrand}, consists of three steps:

\begin{enumerate}
  \item Augmenting the pre-trained branch with a random one to facilitate the learning of new target-specific patterns (\cref{method_add_random});
  
  \item Normalising both branches to balance their behaviours during fine-tuning (\cref{method_norm});
  
  \item Applying learnable weights on both branches to let the network learn which of random or pre-trained one is better for every class. (\cref{method_weighted_vectors}).

\end{enumerate}

%=========================
% Figure PretRand
%=========================

%=========================

%----------------------------
% Adding random branch
%----------------------------
\subsection{Adding the Random Branch}
\label{method_add_random}

We expect that augmenting the pretrained model with new randomly initialised neurons allows a better adaptation during fine-tuning. Thus, in the adaptation stage, we augment the pre-trained model with a random branch consisting of additional random units (as illustrated in the scheme \enquote{a} of Figure \ref{fig:pretrand}). Several works have shown that deep (top) layers are more task-specific than shallow (low) ones \citep{peters2018deep,mou2016transferable}. Thus, deep layers learn generic features easily transferable between tasks. In addition, word embeddings (shallow layers) contain the majority of parameters. Based on these factors, we choose to expand only the top layers as a trade-off between performance and number of parameters (model complexity). In terms of the expanded layers, we add an extra biLSTM layer of $k$ units in the $\mathbf{FE}$ ($\Phi_{r}$ - $r$ for random); and a new fully-connected layer of $C$ units (called $\Psi_{r}$). With this choice, we increase the complexity of the model only $1.02\times$ compared to the base one (The standard fine-tuning scheme).

Concretely, for every $w_i$, two predictions vectors are computed; \(\hat{\mathbf{y}}_{i}^{p}\) from the pre-trained branch and \(\hat{\mathbf{y}}_{i}^{r}\) from the random one. Specifically, the pre-trained branch predicts class-probabilities following:

\begin{equation}
    \hat{\mathbf{y}}^p_{i} = (\Psi_p \circ \Phi_p)(\mathbf{x}_i),
\end{equation}

\noindent
with $\mathbf{x}_i = \Upsilon(w_i)$. Likewise, the additional random branch predicts class-probabilities following:

\begin{equation}
    \hat{\mathbf{y}}^r_{i} = (\Psi_r \circ \Phi_r)(\mathbf{x}_i). 
\end{equation}

\noindent
To get the final predictions, we simply apply an element-wise sum between the outputs of the pre-trained branch and the random branch: 

\begin{equation}
    \hat{\mathbf{y}}_{i} = \hat{\mathbf{y}}_{i}^{p} \oplus \hat{\mathbf{y}}_{i}^{r}. 
\end{equation}

\noindent
As in the classical scheme, the SCE loss is minimised but here, both branches are trained jointly.

%----------------------------------
% Independent normalisation
%----------------------------------
\subsection{Independent Normalisation}
\label{method_norm}

\begin{figure}[ht!]

\begingroup
\setlength{\tabcolsep}{2pt} % Default value: 6pt
\renewcommand{\arraystretch}{0.5} % Default value: 1
\centering
\begin{tabular}{c c}
\includegraphics[scale=0.20]{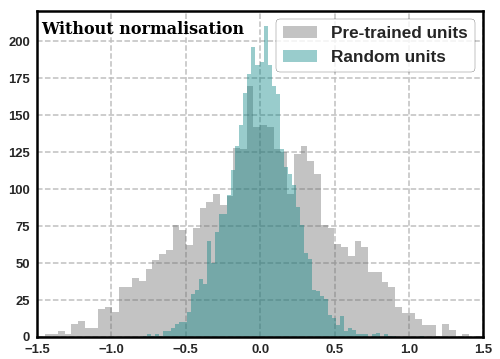}
& \includegraphics[scale=0.20]{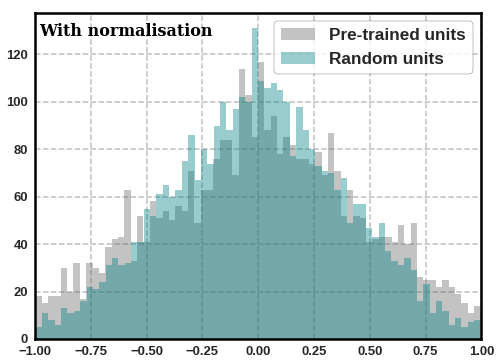}
\end{tabular}

\endgroup

\caption[The distributions of the learnt weight-values for the randomly initialised and pre-trained branches after their joint training.]{The distributions of the learnt weight-values for the randomly initialised (green) and pre-trained (grey) fully-connected layers after their joint training. Left: without normalisation, right: with normalisation. %\textbf{Rajouter les sig des y}
}
\label{fig:FCL_weights}
\end{figure}

Our first implementation of \textit{adding the random branch} was less effective than expected. The main explanation is that the pre-trained units were dominating the random units, which means that the weights as well as the gradients and outputs of pre-trained units absorb those of the random units.  As illustrated in the left plot of Figure \ref{fig:FCL_weights}, the absorption phenomenon stays true even at the end of the training process; we observe that random units weights are closer to zero. This absorption propriety handicaps the random units in firing on the words of the target dataset.\footnote{The same problem was stated in some computer-vision works \citep{liu2015parsenet,wang2017growing,tamaazousti2017mucale}.}

%-------------------------
% Visu: FCL weights without norm
%-------------------------
%-------------------------

To alleviate this absorption phenomenon and push the random units to be more competitive, we normalise the outputs of both branches ($\hat{\mathbf{y}}_{i}^{p}$ and $\hat{\mathbf{y}}_{i}^{r}$) using the \(\ell_2\)-norm, as illustrated in the scheme \enquote{b} of Figure \ref{fig:pretrand}. The normalisation of a vector \enquote{$\mathbf{x}$} is computed using the following formula: 

\begin{equation}
\mathbf{N}_2(\mathbf{x}) = [\frac{\mathbf{x}_i}{||\mathbf{x}||_2}]_{i=1}^{i=|\mathbf{x}|}.
\end{equation}

\noindent
Thanks to this normalisation, the absorption phenomenon was solved, and the random branch starts to be more effective (see the right distribution of Figure \ref{fig:FCL_weights}).

Furthermore, we have observed that despite the normalisation, the performance of the pre-trained classifiers is still much better than the randomly initialised ones. Thus, to make them more competitive, we propose to start with optimising only the randomly initialised units while freezing the pre-trained ones, then, we launch the joint training. We call this technique \textit{random++}.

\subsection{Attention Learnable Weighting Vectors}
\label{method_weighted_vectors}

Heretofore, pre-trained and random branches participate equally for every class' predictions, \textit{i.e.} we do not weight the dimensions of $\hat{\mathbf{y}}_{i}^{p}$ and $\hat{\mathbf{y}}_{i}^{r}$ before merging them with an element-wise summation. 
Nevertheless, random classifiers may be more efficient for specific classes compared to pre-trained ones and vice-versa. In other terms, we do not know which of the two branches (random or pre-trained) is better for making a suitable decision for each class. For instance, if the random branch is more efficient for predicting a particular class $c_j$, it would be better to give more \textit{attention} to its outputs concerning the class $c_j$ compared to the pretrained branch.

Therefore, instead of simply performing an element-wise sum between the random and pre-trained predictions, we first \textit{weight} $\hat{\mathbf{y}}_{i}^{p}$ with a learnable weighting vector $\mathbf{u}~\in~\mathbb{R}^C$ and $\hat{\mathbf{y}}_{i}^{r}$ with a learnable weighting vector $\mathbf{v}~\in~\mathbb{R}^C$, where $C$ is the tagset size (number of classes). Such as, the element $\mathbf{u}_j$ from the vector $\mathbf{u}$ represents the random branch's attention weight for the class $c_j$, and the element $\mathbf{v}_j$ from the vector $\mathbf{v}$ represents the pretrained branch's attention weight for the class $c_j$. Then, we compute a Hadamard product with their associated normalised predictions (see the scheme \enquote{c} of Figure \ref{fig:pretrand}). Both vectors $\mathbf{u}$ and $\mathbf{v}$ are initialised with 1-values and are fine-tuned by back-propagation. 
Formally, the final predictions are computed as follows:

\begin{equation}
\hat{\mathbf{y}}_{i} = \mathbf{u}~\odot~\mathbf{N}_{p}(\hat{\mathbf{y}}_{i}^{p})~\oplus~\mathbf{v}~\odot~\mathbf{N}_{p}(\hat{\mathbf{y}}_{i}^{r}). 
\end{equation}

%=========================
% Experiments
%=========================
\section{Experimental Settings}
\label{sec:experiments}

\definecolor{Gray}{gray}{0.75}
\newcolumntype{g}{>{\columncolor{Gray}}c}

%-------------------------
% Table: datasets statistics 
%-------------------------
\begin{table*}[ht!]
\scriptsize
\centering
\begin{tabular}{l r l l l}
 \hline
 \bf Task & \textbf{\#Classes} & \textbf{Sources} & \textbf{Eval. Metrics} & \textbf{\# Tokens-splits (train - val - test)}\\
 \hline
 POS: POS Tagging               & 36 & WSJ & Top-1 Acc. & 912,344 - 131,768 - 129,654\\
 \hline
 CK: Chunking                   & 22 & CONLL-2000 & Top-1 Acc. & 211,727 - n/a -  47,377\\
 \hline
 NER: Named Entity Recognition  & 4  & CONLL-2003       & Top-1 Exact-match F1. & 203,621 - 51,362 - 46,435 \\

 \hline
 \multirow{3}{*}{MST: Morpho-syntactic Tagging}  & 1304  & Slovene-news       & Top-1 Acc. & 439k - 58k - 88k \\
   \cline{2-5}
 & 772  & Croatian-news       & Top-1 Acc. & 379k - 50k - 75k \\
  \cline{2-5}
 &  557  & Serbian-news       & Top-1 Acc. & 59k - 11k, 16k \\

 \hline
 \hline
 
  \multirow{3}{*}{POS: POS Tagging} & 40 & TPoS & Top-1 Acc. & 10,500 - 2,300 - 2,900\\
  \cline{2-5}
                                & 25 & ArK  & Top-1 Acc. & 26,500 - / - 7,700\\
  \cline{2-5}
                                & 17 & TweeBank & Top-1 Acc. & 24,753 - 11,742 - 19,112\\
 
 \hline
 CK: Chunking                   & 18 & TChunk & Top-1 Top-1 Acc.. & 10,652 -  2,242 -  2,291\\
 \hline
 NER: Named Entity Recognition  & 6  & WNUT-17 & Top-1 Exact-match F1. & 62,729 - 15,734 - 23,394\\
 
 \hline
 \multirow{3}{*}{MST: Morpho-syntactic Tagging}  & 1102  & Slovene-sm       & Top-1 Acc. & 37,756 - 7,056 - 19,296 \\
   \cline{2-5}
 & 654  & Croatian-sm       & Top-1 Acc. & 45,609 - 8,886 - 21,412 \\
  \cline{2-5}
 & 589  & Serbian-sm       & Top-1 Acc. & 45,708- 9,581- 23,327  \\

\hline
\end{tabular}
\caption[Statistics of the used datasets.]{
Statistics of the used datasets. \textbf{Top}: datasets of the source domain. \textbf{Bottom}: datasets of the target domain. }
\label{tab:tasks_and_datasets}
\end{table*}
%-------------------------
%-------------------------

%=========================
% Data Sets
%=========================
\subsection{Datasets}
\label{sec:datasets}

We conduct experiments on supervised domain adaptation from the news domain (formal texts) to the social media  domain (noisy texts) for English Part-Of-Speech tagging (\textbf{POS}), Chunking (\textbf{CK}) and Named Entity Recognition (\textbf{NER}). In addition, we experiment on Morpho-syntactic Tagging (\textbf{MST}) of three South-Slavic languages: Slovene, Croatian and Serbian. For POS task, we use the WSJ part of Penn-Tree-Bank (PTB) \citep{marcus1993building} news dataset for the source news domain and  TPoS \citep{ritter2011named}, ArK \citep{owoputi2013improved} and TweeBank \citep{liu2018parsing} for the target social media domain. For CK task, we use the CONLL2000 \citep{tjong2000introduction} dataset for the news source domain and TChunk \citep{ritter2011named} for the target domain. For NER task, we use the CONLL2003 dataset \citep{tjong2003introduction} for the source news domain and WNUT-17 dataset \citep{derczynski2017results} for the social media target domain. For MST, we use the MTT shared-task \citep{zampieri2018language} benchmark containing two types of datasets: social media and news, for three south-Slavic languages: Slovene (sl), Croatian (hr) and Serbian (sr). Statistics of all the datasets are summarised in Table \ref{tab:tasks_and_datasets}. %More details about the tasks and datasets are provided in Appendix \ref{sec:appendix_datasets}.

%=========================
% Data Sets
%=========================
\subsection{Evaluation Metrics}
\label{sec:eval} 

We evaluate our models using metrics that are commonly used by the community. Specifically, accuracy (acc.) for POS, MST and CK and entity-level F1 for NER.

\textbf{Comparison criteria}: A common approach to compare the performance between different approaches across different datasets and tasks is to take the average of each approach across all tasks and datasets. However, as it has been discussed in many research papers \citep{subramanian2018learning,rebuffi2017learning,tamaazousti2018universality}, when tasks are not evaluated using the same metrics or results across datasets are not of the same order of magnitude, the simple average does not allow a \enquote{coherent aggregation}. For this, we use the average Normalized Relative Gain (aNRG) proposed by \citet{tamaazousti2018universal}, where a score $\mathbf{aNRG_{i}}$ for each approach $i$ is calculated compared to a reference approach (baseline) as follows:

\begin{equation}
\label{eqn:anrg}
    \mathbf{aNRG_{i}}~=~\frac{1}{L}~\sum_{j=1}^{L}~\frac{(s_j^{i}~-~s_j^{ref})}{(s_j^{max}~-~s_j^{ref})}~,
\end{equation}

\noindent
with $s^{i}_{j}$ being the score of the approach$_i$ on the dataset$_j$, $s^{ref}_{j}$ being the score of the reference approach on the dataset$_j$ and $s^{max}_{j}$ is the best achieved score across all approaches on the dataset$_j$.

\subsection{Implementation Details}
\label{section_experimental_setup}

\noindent
We use the following Hyper-Parameters (HP):

\noindent
\textbf{WRE's HP}: In the standard word-level embeddings, tokens are lower-cased while the character-level component still retains access to the capitalisation information. We set the randomly initialised character embedding dimension at 50, the dimension of hidden states of the character-level biLSTM at 100 and used 300-dimensional word-level embeddings. The latter were pre-loaded from publicly available GloVe pre-trained vectors on 42 billions words from a web crawling and containing 1.9M words \citep{pennington2014glove} for English experiments, and pre-loaded from publicly available FastText \citep{bojanowski2017enriching} pre-trained vectors on common crawl for South-Slavic languages.\footnote{\url{https://github.com/facebookresearch/fastText/blob/master/docs/crawl-vectors.md}} These embeddings are also updated during training. For experiments with contextual words embeddings (\cref{sec:results_plus_elmo}), we used ELMo (Embeddings from Language Models) embeddings~\citep{peters2018deep}. For English, we use the small official pre-trained ELMo model on 1 billion word benchmark (13.6M parameters).\footnote{\url{https://allennlp.org/elmo}} Regarding South-Slavic languages, ELMo pre-trained models are not available but for Croatian \citep{che2018towards}.\footnote{\url{https://github.com/HIT-SCIR/ELMoForManyLangs}} Note that, in all experiments contextual embeddings are frozen during training.

\noindent
\textbf{FE's HP}: we use a single biLSTM layer (token-level feature extractor) and set the number of units to 200.

\noindent
\textbf{PretRand's random branch HP}: we experiment our approach with $k=200$ added random-units. 

\noindent
\textbf{Global HP}: In all experiments, training (pretraining and fine-tuning) are performed using the SGD with momentum with early stopping, mini-batches of 16 sentences and learning rate of $1.5\times10^{-2}$. All our models are implemented with the PyTorch library \citep{paszke2017automatic}.

%=========================
% Results
%=========================
\section{Experimental Results}
\label{sec:results}

This section reports all our experimental results and analysis. First we analyse the standard fine-tuning scheme of transfer learning (\cref{sec:analysis_sft_results}). Then we assess the performance of our proposed approach, PretRand (\cref{sec:pretrand_results}).

%=========================
% Results
%=========================
\subsection{Analysis of the Standard Fine-tuning Scheme}
\label{sec:analysis_sft_results}

We report in Table~\ref{table:main_results} the results of the reference \textit{supervised training scheme from scratch}, followed by the results of the standard fine-tuning scheme, which outperforms the reference. Precisely, transfer learning exhibits an improvement of $\sim$+3\% acc. for TPoS, $\sim$+1.2\% acc. for ArK, $\sim$+1.6\% acc. for TweeBank, $\sim$+3.4\% acc. for TChunk and $\sim$+4.5\% F1 for WNUT.

%%%%%% Table : main results  %%%%%%%
\begin{table*} [h!]

\small
\centering
\begin{tabular}{|l|c c|c|c c|c c|c|}

\cline{2-9}
\multicolumn{1}{c|}{} & \multicolumn{5}{c|}{POS (Acc.)} &  \multicolumn{2}{ c|}{CK (Acc.)} & NER (F1) \\ \hline  
\multirow{2}{*}{\diagbox[innerrightsep=12pt]{\bf Method}{\bf Dataset}} & \multicolumn{2}{ c|}{TPoS} & ARK & \multicolumn{2}{ c|}{Tweebank} & \multicolumn{2}{ c|}{TChunk} & WNUT \\
 & dev & test & test & dev & test & dev & test & test \\  \hline

\bf From scratch & 88.52 & 86.82 & 90.89 & 91.61 & 91.66 & 87.76 & 85.83 & 36.75  \\
\bf Standard Fine-tuning &  \textbf{90.95} & \textbf{89.79} & \textbf{92.09} & \textbf{93.04} & \textbf{93.29} & \textbf{90.71} & \textbf{89.21} & \textbf{41.25} \\
\hline
\end{tabular}
\caption[Main results of our proposed approach, \textit{transferring pretrained models}. ]{The main results of our proposed approach, \textit{transferring pretrained models}, on social media datasets (Acc (\%) for POS and CK and F1 (\%) for NER). The best score for each dataset is highlighted in bold.}
\label{table:main_results} 

\end{table*}

%%%%%%%%%%%%%%%%%%%%%%%%%%%%

In the following we provide the results of our analysis of the standard fine-tuning scheme:

\begin{enumerate}
    \item Analysis of the hidden negative transfer (\cref{sec:proposed_method_negative_transfer_results}).
    
    \item Quantifying the change of individual pretrained neurons after fine-tuning (\cref{sec:proposed_visualization_correlation_results}).
    
    \item Visualising the evolution of pretrained neurons stimulus during fine-tuning (\cref{sec:proposed_visualization_units_topk_results}).
\end{enumerate}

%=========================
% Analysis of the Hidden Negative Transfer
%=========================
\subsubsection{Analysis of the Hidden Negative Transfer}
\label{sec:proposed_method_negative_transfer_results} 

To investigate the \textit{hidden} negative transfer in the standard fine-tuning scheme of transfer learning, we propose the following experiments. First, we show that the final gain brought by the standard fine-tuning can be separated into two categories: \textit{positive transfer} and \textit{negative transfer}. Second, we provide some qualitative examples of negative transfer. \\

%%%%%%%%%%%%%%%%%%%%%%%%%%%%%%%%%
%%% Quantifying Negative Transfer
%%%%%%%%%%%%%%%%%%%%%%%%%%%%%%%%%

\noindent
\underline{{Quantifying Positive Transfer \& Negative Transfer}}
%\label{sec:quantifying_negative_transfer}

%-------------------------
% Visu: Quantification of Negative Transfer
%-------------------------
\begin{figure}[ht!]
\centerline{\includegraphics[scale=0.30]{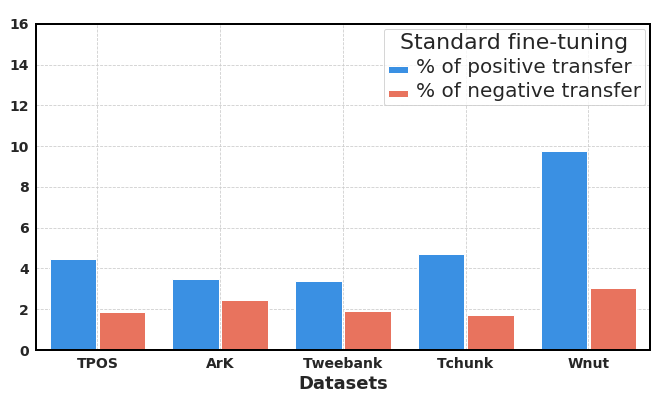}}
\caption[The percentage of \textit{negative transfer} and positive transfer brought by the standard fine-tuning adaptation scheme compared to supervised training from scratch.]{The percentage of negative transfer and positive transfer brought by the standard fine-tuning adaptation scheme compared to supervised training from scratch scheme.
}
\label{fig:negative_transfer}
\end{figure}
%-------------------------
%-------------------------

\noindent

We recall that we define \textit{positive transfer} as the percentage of tokens that were wrongly predicted by random initialisation (supervised training from scratch), but the standard fine-tuning changed to the correct ones, while \textit{negative transfer} represents the percentage of words that were tagged correctly by random initialisation, but using standard fine-tuning gives wrong predictions. Figure \ref{fig:negative_transfer} shows the results on English social media datasets, first tagged with the classic supervised training scheme and then using the standard fine-tuning. Blue bars show the percentage of \textit{positive transfer} and red bars give the percentage of \textit{negative transfer}. We observe that even though the standard fine-tuning approach is effective since the resulting \textit{positive transfer} is higher than the \textit{negative transfer} in all cases, this last mitigates the final gain brought by the standard fine-tuning. For instance, for TChunk dataset, standard fine-tuning corrected $\sim$4.7\% of predictions but falsified $\sim$1.7\%, which reduces the final gain to $\sim$3\%.\footnote{Here we calculate positive and negative transfer at the token-level. Thus, the gain shown in Figure \ref{fig:negative_transfer} for WNUT dataset does not correspond to the one in Table \ref{table:main_results}, since the F1 metric is calculated only on named-entities.}\\

%-------------------------
% Table: Falsified predictions by SFT
%-------------------------
\begin{table*}[ht!]

\small
\centering
  \begin{tabular}{l c c c c c c c}
 \hline

 \bf DataSet & & & & & & & \\
 
 \hline

%%%%%%%%%%%%%%%%%%%%%%
 TPoS  & Award$^{\diamond}$ & 's & its$^{\star}$ & Mum & wont$^{\star}$ & id$^{\star}$ & Exactly \\

           &  \textcolor{teal}{nn} & \textcolor{teal}{vbz}   &  \textcolor{teal}{prp} &  \textcolor{teal}{nn}    &  \textcolor{teal}{MD}    &  \textcolor{teal}{prp} &  \textcolor{teal}{uh} \\  
           
           & \textcolor{red}{nnp} & \textcolor{red}{pos} &  \textcolor{red}{prp\$} & \textcolor{red}{uh} & \textcolor{red}{VBP}  &  \textcolor{red}{nn} & \textcolor{red}{rb}  \\

 \hline
 
%%%%%%%%%%%%%%%%%%%%%%
 
 ArK  & Charity$^{\diamond}$ & I'M$^{\star}$ & 2pac$^{\times}$ & 2$^{\times}$ & Titans$^{\star}$ & wth$^{\times}$ & nvr$^{\times}$ \\
 
          & \textcolor{teal}{noun} &  \textcolor{teal}{L}  &  \textcolor{teal}{pnoun} &  \textcolor{teal}{P} &  \textcolor{teal}{Z}  &  \textcolor{teal}{!}    &  \textcolor{teal}{R}   \\  
          
          & \textcolor{red}{pnoun}  & \textcolor{red}{E} & \textcolor{red}{\$}   & \textcolor{red}{\$} & \textcolor{red}{N}  & \textcolor{red}{P}   & \textcolor{red}{V}   \\
          
 \hline

%%%%%%%%%%%%%%%%%%%%%%
TweeBank  & amazin$^{\bullet}$ & Night$^{\diamond}$ & Angry$^{\diamond}$ & stangs & \#Trump & awsome$^{\bullet}$ & bout$^{\bullet}$ \\
          
          & \textcolor{teal}{adj} &  \textcolor{teal}{noun} & \textcolor{teal}{adj} &  \textcolor{teal}{propn} &  \textcolor{teal}{propn} &  \textcolor{teal}{adj}    &  \textcolor{teal}{adp} \\  
          
          & \textcolor{red}{noun}  & \textcolor{red}{propn} & \textcolor{red}{propn} & \textcolor{red}{noun}  & \textcolor{red}{X} & \textcolor{red}{intj}   & \textcolor{red}{verb} \\
          
 \hline 

%%%%%%%%%%%%%%%%%%%%%%

TChunk & luv$^{\times}$ & \multicolumn{2}{c}{**ROCKSTAR**THURSDAY}  & ONLY & Just$^{\diamond}$ & wyd$^{\times}$ & id$^{\star}$\\

    & \textcolor{teal}{b-vp}  &  \multicolumn{2}{c}{\textcolor{teal}{b-np}} &  \textcolor{teal}{i-np}   &  \textcolor{teal}{b-advp} & \textcolor{teal}{b-np} & \textcolor{teal}{b-np}\\  
 
    & \textcolor{red}{i-intj}  & \multicolumn{2}{c}{\textcolor{red}{O}} & \textcolor{red}{b-np}  & \textcolor{red}{b-np}   & \textcolor{red}{b-intj} & \textcolor{red}{i-np}\\

 \hline

%%%%%%%%%%%%%%%%%%%%%%

Wnut  & Hey$^{\diamond}$ & Father$^{\diamond}$ & \&$^{\times}$ & IMO$^{\times}$ & UN & Glasgow & Supreme \\

          & \textcolor{teal}{O} &  \textcolor{teal}{O}  &  \textcolor{teal}{O}   &  \textcolor{teal}{O}   &  \textcolor{teal}{O}  &  \textcolor{teal}{b-location}    &  \textcolor{teal}{b-person} \\  
          
          & \textcolor{red}{b-person}  & \textcolor{red}{b-person} & \textcolor{red}{i-group}  & \textcolor{red}{b-group}  & \textcolor{red}{b-group}  & \textcolor{red}{b-group}   & \textcolor{red}{b-corporation} \\

\hline

%%%%%%%%%%%%%%%%%%%%%%
\end{tabular}
\begin{flushleft}
\scriptsize{nn=N=noun=common noun / nnp=pnoun=propn=proper noun / vbz=Verb, 3rd person singular present / pos=possessive ending / prp=personal pronoun / prp\$=possessive pronoun / md=modal / VBP=Verb, non-3rd person singular present / uh=!=intj=interjection / rb=R=adverb / L=nominal + verbal or verbal + nominal / E=emoticon / \$=numerical / P=pre- or postposition, or subordinating conjunction / Z=proper noun + possessive ending / V=verb / adj=adjective / adp=adposition}
\end{flushleft}
\caption[Examples of falsified predictions by the standard fine-tuning scheme of transfer learning.]{\textbf{Examples of falsified predictions by standard fine-tuning scheme when transferring from news domain to social media domain}. Line 1: Some words from the validation-set of each data-set. \textcolor{teal}{Line 2: Correct labels predicted by the classic supervised setting (Random-200)}. \textcolor{red}{Line 3: Wrong labels predicted by SFT setting}. Mistake type: $\diamond$ for words with first capital letter, $\bullet$ for misspelling, $\star$ for contractions, $\times$ for abbreviations.}
\label{tab:examples}

\end{table*}

%%%%%%%%%%%%%%%%%%%%%%
% Visu: Units correlation
%%%%%%%%%%%%%%%%%%%%%%
\begin{figure*}
\begin{tabular}{c c c }
\centering
\includegraphics[scale=0.22]{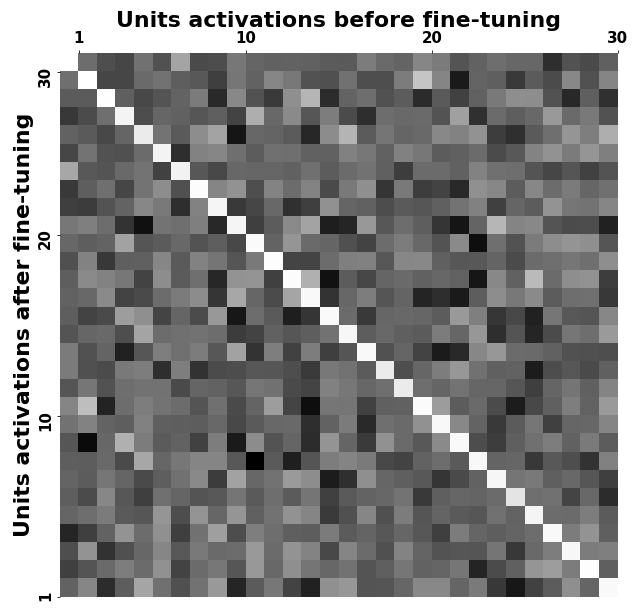}
&
\includegraphics[scale=0.22]{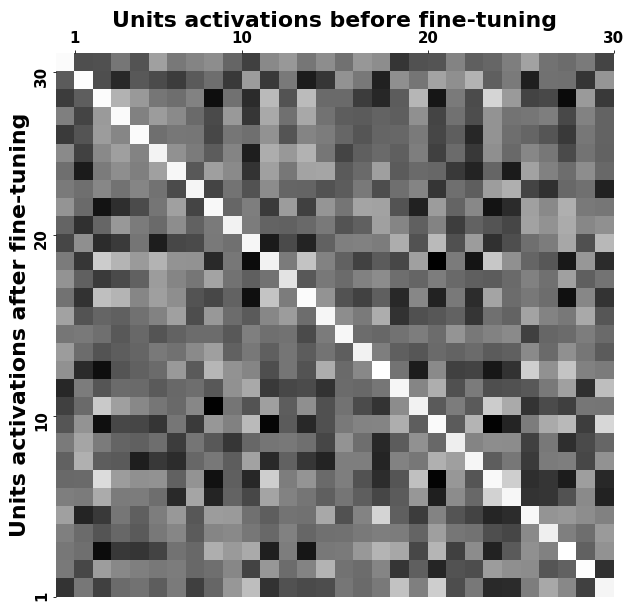}
&
\includegraphics[scale=0.22]{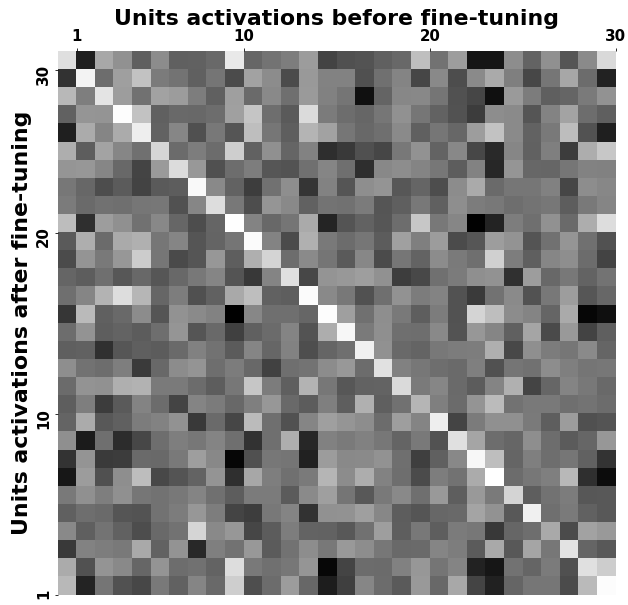}
\end{tabular}

\begin{tablenotes}
 \item \qquad   \qquad \qquad \scriptsize{\bf ArK dataset} \qquad  \quad \qquad \qquad  \qquad \qquad \qquad \qquad \scriptsize{\bf Tchunk dataset} \qquad \qquad  \qquad \qquad \qquad \qquad \qquad \scriptsize{\bf Wnut dataset}
\end{tablenotes}

\caption[Correlation results between $\Phi$ units' activations before fine-tuning and after fine-tuning.]{Correlation results between $\Phi$ units' activations before fine-tuning (columns) and after fine-tuning (rows). Brighter colours indicate higher correlation.}
\label{fig:units_corr}
\end{figure*}
%%%%%%%%%%%%%%%%%%%%%%

%=========================
% Qualitative examples of negative transfer
%=========================
\noindent
\underline{{Qualitative Examples of Negative Transfer}}
%\label{sec:analysis_bias_sft_qualitative}

\noindent
We report in Table \ref{tab:examples} concrete examples of words whose predictions were falsified when using the standard fine-tuning scheme compared to standard supervised training scheme. Among mistakes we have observed:

\begin{itemize}

    \item \textbf{Tokens with an upper-cased first letter}: In news (formal English), only proper nouns start with an upper-case letter inside sentences. Consequently, when using transfer learning, the pre-trained units fail to slough this pattern which is not always respected in social media. Hence, we found that most of the tokens with an upper-cased first letter are mistakenly predicted as proper nouns (PROPN) in POS, \textit{e.g.} Award, Charity, Night, etc. and as entities in NER, \textit{e.g.} Father, Hey, etc., which is consistent with the findings of \citet{seah2012combating}: negative transfer is mainly due to conditional distribution differences between source and target domains.
    
    \item \textbf{Contractions} are frequently used in social media to shorten a set of words. For instance, in TPoS dataset, we found that \enquote{'s} is in most cases predicted as a \enquote{possessive ending (pos)} instead of \enquote{Verb, 3rd person singular present (vbz)}. Indeed, in formal English, \enquote{'s} is used in most cases to express the possessive form, \textit{e.g.} \enquote{company's decision}, but rarely in contractions that are frequently used in social media, \textit{e.g.} \enquote{How's it going with you?}. Similarly, \enquote{wont}  is a frequent contraction for \enquote{will not}, \textit{e.g.} \enquote{i wont get bday money lool}, predicted as \enquote{verb} instead of \enquote{modal (MD)}\footnote{A modal is an auxiliary verb expressing: ability (can), obligation (have), etc.} by the SFT scheme. The same for \enquote{id}, which stands for \enquote{I would}.
    
    \item \textbf{Abbreviations} are frequently used in social media to shorten the way a word is standardly written. We found that the standard fine-tuning scheme stumbles on abbreviations predictions, \textit{e.g.} \textit{2pac} (Tupac), \textit{2} (to), \textit{ur} (your), \textit{wth} (what the hell) and \textit{nvr} (never) in ArK dataset; and \textit{luv} (love) and \textit{wyd} (what you doing?) in TChunk dataset.

    \item \textbf{Misspellings}: Likewise, we found that the standard fine-tuning scheme often gives wrong predictions for misspelt words, \textit{e.g.} \textit{awsome}, \textit{bout}, \textit{amazin}.

\end{itemize}

%=========================
% Sec: Correlation Analysis
%=========================
\subsubsection{Quantifying the change of individual pretrained neurons}
\label{sec:proposed_visualization_correlation_results}

To visualise the bias phenomenon occurring when using the standard fine-tuning scheme, we quantify the \textit{charge} of individual neurons. Precisely, we plot the asymmetric correlation matrix $\mathbf{C}$ (The method described in \cref{sec:proposed_visualization_correlation}) between the $\Phi$ layer's units \textit{before} and \textit{after} fine-tuning for each social media dataset (ArK for POS, TChunk for CK and WNUT-17 for NER). From the resulting correlation matrices illustrated in Figure \ref{fig:units_corr}, we can observe the diagonal representing the \textit{charge} of each unit, with most of the units having a high charge (light colour), alluding the fact that every unit after fine-tuning is highly correlated with itself before fine-tuning. Hypothesising that high correlation in the diagonal entails high bias, the results of this experiment confirm our initial motivation that pre-trained units are highly biased to what they have learnt in the source-dataset, making them limited to learn some patterns that are specific to the target-dataset. Our remarks were confirmed recently in the recent work of \citet{merchant2020hapepns} who also found that fine-tuning is a \enquote{conservative process}.

%=========================
% Sec: Individual units visualisation during adaptation
%=========================
\subsubsection{Visualising the Evolution of Pretrained Neurons Stimulus during Fine-tuning}
\label{sec:proposed_visualization_units_topk_results}

Here, we give concrete visualisations of the evolution of pretrained neurons stimulus during fine-tuning when transferring from the news domain to the social media domain. Following the method described in section \ref{sec:proposed_visualization_units_topk}, we plot the matrices of top-10 words activating each neuron $j$, positively ($\mathbf{A}^{(j)}_{best+}$) or negatively ($\mathbf{A}^{(j)}_{best-}$). The results are plotted in Figure \ref{fig:neurons_sft_top_act_ark} for ArK (POS) dataset and Figure \ref{fig:neurons_sft_top_act_tweebank} for TweeBank dataset (POS). Rows represent the top-10 words from the target dataset activating each unit, and columns represent fine-tuning epochs; before fine-tuning in column 0 (at this stage the model is only trained on the source-dataset), and during fine-tuning (columns 5 to 20). Additionally, to get an idea about each unit's stimulus on source dataset, we also show, in the first column (Final-WSJ), top-10 words from the source dataset activating the same unit before fine-tuning. In the following, we describe the information encoded by each provided neuron.\footnote{Here we only select some interesting neurons. However we also found many neurons that are not interpretable.}

\begin{itemize}

    \item  \textbf{Ark - POS:} (Figure \ref{fig:neurons_sft_top_act_ark})
        
        \begin{itemize}

        \item Unit-196 is sensitive to contractions containing an apostrophe regardless of the contraction's class. However, unlike news, in social media and particularly ArK dataset, apostrophes are used in different cases. For instance \textit{i'm}, \textit{i'll} and \textit{it's} belong to the class \enquote{L} that stands for \enquote{nominal + verbal or  verbal + nominal}, while the contractions \textit{can't} and \textit{don't} belong to the class \enquote{Verb}. 
        
        \item Unit-64 is sensitive to plural proper nouns on news-domain before fine-tuning, \textit{e.g.} \textit{Koreans} and \textit{Europeans}, and also on ArK during fine-tuning, \textit{e.g.} \textit{Titans} and \textit{Patriots}. However, in ArK dataset, \enquote{Z} is a special class for \enquote{proper noun + possessive ending}, \textit{e.g.} \textit{Jay's mum}, and in some cases the apostrophe is omitted, \textit{e.g.} \textit{Fergusons house} for \textit{Ferguson's house}, which thus may bring ambiguity with plural proper nouns in formal English. Consequently, unit-64, initially sensitive to plural proper nouns, is also firing on words from the class \enquote{Z}, e.g. \textit{Timbers} (\textit{Timber's}).

        \end{itemize}

%%%%%%%%%%%%%%%%%%%%%%
% Visu: Units from SFT scheme => top-K w.r.t epoch
%%%%%%%%%%%%%%%%%%%%%%
\begin{figure}

%%%%%% ARK   %%%%%%%%
%{\begin{tabnote}
%\textbf{\textcolor{pos_green_2}{ArK dataset}}
\begin{tablenotes}
\item \scriptsize{\bf Unit-196: ArK dataset}
\end{tablenotes}

\centering
\begin{tabular}{c}
\includegraphics[scale=0.25]{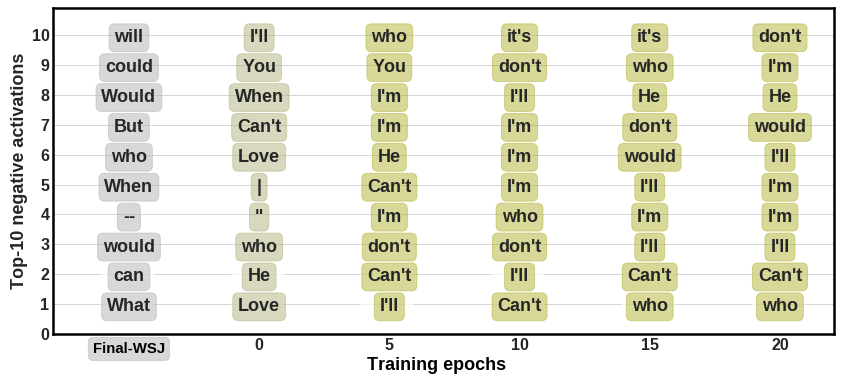}
\end{tabular}

\begin{tablenotes}
\item \scriptsize{\bf Unit-64: ArK dataset}
\end{tablenotes}

\centering
\begin{tabular}{c}
\includegraphics[scale=0.25]{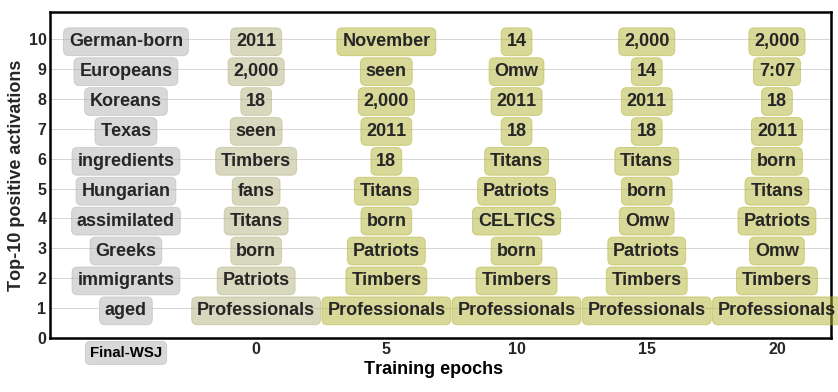}
\end{tabular}

%%%%%%%%%%%%%%%%%%%%%%

\caption[Individual units activations before and during fine-tuning from ArK POS dataset.]{\textbf{Individual units activations before and during fine-tuning from ArK POS dataset}. For each unit we show Top-10 words activating the said unit. The first column: top-10 words from the source validation-set (WSJ) before fine-tuning, Column 0: top-10 words from the target validation-set (ArK) before fine-tuning. Columns 5 to 20: top-10 words from the target validation-set during fine-tuning epochs.
}
\label{fig:neurons_sft_top_act_ark}
\end{figure}
%%%%%%%%%%%%%%%%%%%%%%
%%%%%%%%%%%%%%%%%%%%%%

        \item \textbf{Tweebank - POS:} (Figure \ref{fig:neurons_sft_top_act_tweebank})
        
        \begin{itemize}

        \item Unit-37 is sensitive before and during fine-tuning on plural nouns, such as \textit{gazers} and \textit{feminists}. However, it is also firing on the word \textit{slangs} because of the \textit{s} ending, which is in fact a proper noun. This might explain the wrong prediction for the word \textit{slangs} (noun instead of proper noun) given by the standard fine-tuning scheme (Table \ref{tab:examples}). 
        
        \item Unit-169 is highly sensitive to proper nouns (\textit{e.g.} \textit{George} and \textit{Washington}) before fine-tuning, and to words with capitalised first-letter whether the word is a proper noun or not (\textit{e.g.} \textit{Man} and \textit{Father}) during fine-tuning on the TweeBank dataset. Which may explain the frequent wrong predictions of tokens with upper-cased first letter as proper nouns by the standard fine-tuning scheme.
        
        \end{itemize}

%%%%%%%%%%%%%%%%%%%%%%
% Visu: Units from SFT scheme => top-K w.r.t epoch
%%%%%%%%%%%%%%%%%%%%%%
\begin{figure}[ht!]
%%%%%% Tweebank dataset %%%%%%%%
\begin{tablenotes}
\item \scriptsize{\bf Unit-37: Tweebank dataset}
\end{tablenotes}
\centering
\begin{tabular}{c}
\centerline{\includegraphics[scale=0.25]{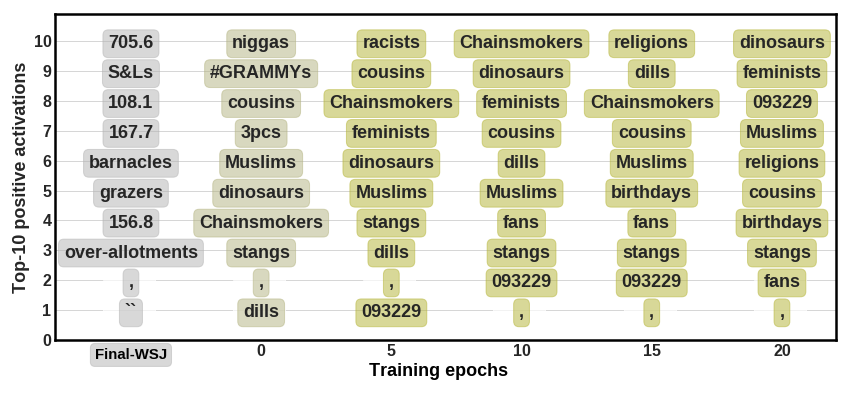}}
\end{tabular}
\begin{tablenotes}
\item \scriptsize{\bf Unit-169: Tweebank dataset}
\end{tablenotes}
\centering
\begin{tabular}{c}
\centering
\centerline{\includegraphics[scale=0.25]{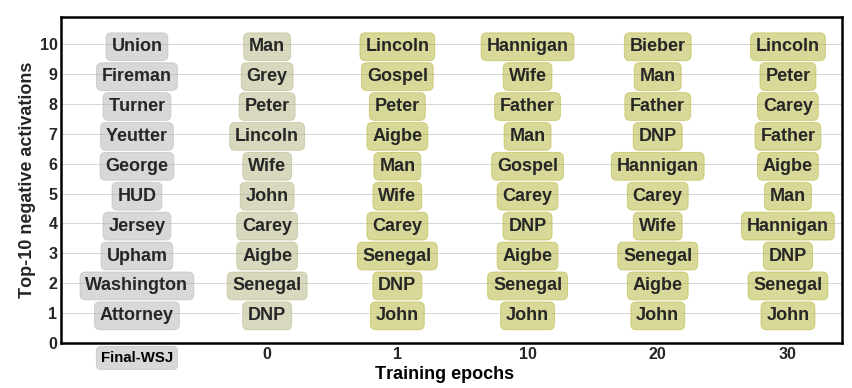}}
\end{tabular}
\caption[Individual units activations before and during fine-tuning on Tweebank POS dataset.]{\textbf{Individual units activations before and during fine-tuning on Tweebank POS dataset}. For each unit we show Top-10 words activating the said unit. The first column: top-10 words from the source validation-set (WSJ) before fine-tuning, Column 0: top-10 words from the target validation-set (Tweebank) before fine-tuning. Columns 5 to 20: top-10 words from the target validation-set during fine-tuning epochs.
}
\label{fig:neurons_sft_top_act_tweebank}
\end{figure}

\end{itemize}

%=========================
% Results
%=========================
\subsection{PretRand's Results}
\label{sec:pretrand_results}

In this section, we present PretRand's performance on POS, CK, NER and MST tasks on social media datasets:

\begin{enumerate}
    \item We compare PretRand's to baseline methods, in the scenario in which contextual representations (ELMo) are not used (\cref{results_compare_to_baselines}).
    
    \item We measure the importance of each component of PretRand on the overall performance (\cref{sec:pretrand_results_ablation_study}).
    
    \item We investigate the impact of incorporating contextual representations, on baselines \textit{vs} PretRand (\cref{sec:results_plus_elmo}).
    
    \item We compare PretRand to best state-of-the-art approaches (\cref{sec:results_compare_with_SOTA}).
    
    \item We investigate in which scenarios PretRand is most advantageous (\cref{sec:pretrand_when_where}).
    
    \item We assess the impact of PretRand on the hidden negative transfer compared to the standard fine-tuning (\cref{sec:analysis_pretrand_negative_transfer})
    
\end{enumerate}

%=========================
% Results: Comparison with Baseline Methods
%=========================
\subsubsection{Comparison with Baseline Methods}
\label{results_compare_to_baselines}

In this section we assess the performance of PretRand through a comparison to six baseline-methods, illustrated in Figure \ref{fig:baselines_schmes_pretrand}. First, since PretRand is an amelioration of the standard fine-tuning (SFT) adaptation scheme, we mainly compare it to the SFT baseline. Besides, we assess whether the gain brought by PretRand is due to the increase in the number of parameters; thus we also compare with the standard supervised training scheme with a wider model. Finally, the final predictions of PretRand are the combination of the predictions of the two branches, randomly initialised and pretrained, which can make one think about ensemble methods \citep{dietterich2000ensemble}. Thus we also compare with ensemble methods. The following items describe the different baseline-methods used for comparison:

%-------------------------
% Illustrative schemes of baseline-methods and PretRand
%-------------------------
\begin{figure*}
    \centering
    \includegraphics[scale=0.036]{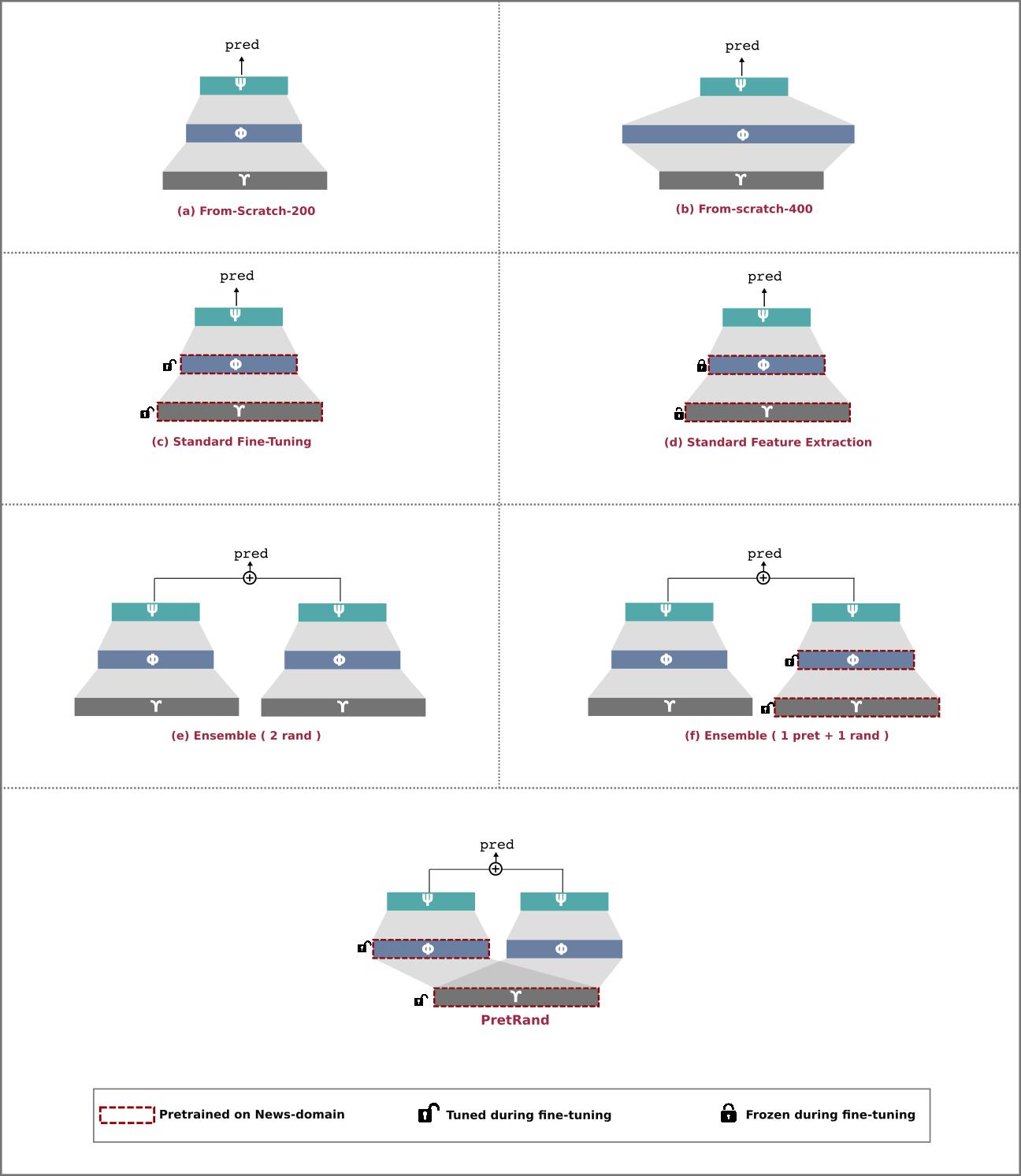}
    \caption{Illustrative schemes of baseline-methods and PretRand.  }
\label{fig:baselines_schmes_pretrand}
\end{figure*}
%---------------------------

\begin{itemize}

    \item \textbf{(a) From-scratch$_{200}$}: The base model described in section \ref{fig:base_architecture}, trained from scratch using the standard supervised training scheme on social media dataset (without transfer learning). Here the number 200 refers to the dimensionality of the biLSTM network in the $\mathbf{FE}$ ($\Phi$).
    
    \item \textbf{(b) From-scratch$_{400}$}: The same as \enquote{From-scratch$_{200}$} baseline but with 400 instead of 200 biLSTM units in the $\mathbf{FE}$. Indeed, by experimenting with this baseline, we aim to highlight that the impact of PretRand is not due to the increase in the number of parameters.

    \item \textbf{(c) Standard Fine-tuning (SFT)}: Pre-training the base model on the source-dataset, followed by an adaptation on the target-dataset with the standard fine-tuning scheme (\cref{sec:sft_scheme}).
    
    \item \textbf{(d) Standard Feature Extraction (SFE)}: The same as SFT, but the pretrained parameters are frozen during fine-tuning on the social media datasets.

    \item \textbf{(e) Ensemble (2 rand)}: Averaging the predictions of two base models that are randomly initialised and learnt independently on the same target dataset, but with a different random initialisation.  
    
    \item \textbf{(f) Ensemble (1 pret + 1 rand)}: same as the previous but with one pre-trained on the source-domain (SFT baseline) and the other randomly initialised (From-scratch$_{200}$ baseline).

\end{itemize}

%%%%%%%%%%%%%%%%%%%%%%
% Table: Perfs on  all datasets
%%%%%%%%%%%%%%%%%%%%%%

\definecolor{Gray2}{gray}{0.9}
\newcolumntype{H}{>{\setbox0=\hbox\bgroup}c<{\egroup}@{}}

\begin{table*}[ht!]

\small
\centering

\begin{tabular}{|l | l | c c c | c |  c H H H | c|}
 \hline

 \multirow{ 2}{*}{\bf Method} & \multirow{ 2}{*}{\textbf{\#params}} & \multicolumn{3}{ c|}{\textbf{POS (acc.)}} & \multicolumn{1}{c|}{\textbf{CK (acc.)}} & \multicolumn{1}{c}{\textbf{NER (F1)}} & & & & \multirow{ 2}{*}{\textcolor{Gray}{aNRG}}\\ 

  &   &  \bf TPoS &  \bf ArK &  \bf TweeBank &  \bf TChunk &   \bf WNUT & \bf Serbian & \bf Slovene & \bf Croatian &\\

\hline
%\rowcolor{Gray2}

\bf From-scratch$_{200}$  & $1\times$ & 86.82 & 91.10 & 91.66 & 85.96 & 36.75 &  86.18 &  84.42  & 85.67 & \textcolor{Gray}{0}\\

\bf From-scratch$_{400}$  & $1.03\times$ & 86.61 &  91.31 &   91.81 & 87.11 & 38.64 &  86.05 & 84.37  & 85.77 & \textcolor{Gray}{+2.7}\\

\cdashline{1-11}
\bf Feature Extraction & $1\times$ & 86.08 & 85.25 & 87.93 & 81.49 & 27.83 & 73.56 &  70.22  & 79.11 &\textcolor{Gray}{-32.4}\\ 

\bf Fine-Tuning & $1\times$ & \underline{89.57} & \underline{92.09} &  \underline{93.23} & \underline{88.86} & 41.25 &  87.59 & \underline{88.76}  & 88.79&\textcolor{Gray}{+15.7}\\

\cdashline{1-11}
\bf Ensemble (2 rand)& $2\times$  &   88.98   &  91.45  & 92.26 &  86.72  & 39.54 & 87.01 &  84.67  & 86.05 &\textcolor{Gray}{+7.5}\\

\bf Ensemble (1p+1r)& $2\times$   & 88.74  & 91.67 & 93.06 & 88.78 &  \underline{42.66}  
& \underline{87.96} & 88.54  & \underline{88.87} &\textcolor{Gray}{+13.4}\\

\cdashline{1-11}
\bf PretRand & $1.02\times$  & \textbf{91.27}  & \textbf{93.81} & \textbf{95.11}  &  \textbf{89.95}  & \textbf{43.12} &  \textbf{88.21} & \textbf{90.01}  & \textbf{90.23} &\textcolor{Gray}{+28.8}\\

\hline
\end{tabular}

\vspace{+0.5cm}
\begin{tabular}{|l | l | H H H H H c c c | c |}
\hline
\multirow{ 2}{*}{\bf Method} & \multirow{ 2}{*}{\textbf{\#params}} & & & & & & \multicolumn{3}{ c|}{\textbf{MST (acc.)}} & \multirow{ 2}{*}{\textcolor{Gray}{aNRG}} \\ 
&   &  \bf TPoS &  \bf ArK &  \bf TweeBank &  \bf TChunk &   \bf WNUT & \bf Serbian & \bf Slovene & \bf Croatian & \\
\hline

%\rowcolor{Gray2}

\bf From-scratch$_{200}$ & $1\times$ & 86.82 & 91.10 & 91.66 & 85.96 & 40.36 &  86.18 &  84.42  & 85.67 & \textcolor{Gray}{0}\\ 

\bf From-scratch$_{400}$ & $1.03\times$ & 86.61 &  91.31 &   91.81 & 87.11 & 42.28 &  86.05 & 84.37  & 85.77 & \textcolor{Gray}{-0.2}\\

\cdashline{1-11}
\bf Feature Extraction & $1\times$ & 86.08 & 85.25 & 87.93 & 81.49 & 37.78 & 73.56 &  70.22  & 79.11 & \textcolor{Gray}{-76.1}\\

\bf Fine-Tuning & $1\times$ & \underline{89.57} & \underline{92.09} &  \underline{93.23} & \underline{88.86} & 41.92 &  87.59 & \underline{88.76}  & 88.79 & \textcolor{Gray}{+19.9}\\
\cdashline{1-11}
\bf Ensemble (2 rand)& $2\times$  &   88.98   &  91.45  & 92.26 &  86.72  & 41.79 & 87.01 &  84.67  & 86.05 & \textcolor{Gray}{+3.4}\\

\bf Ensemble (1p+1r)& $2\times$   & 88.74  & 91.67 & 93.06 & 88.78 &  \underline{42.25}  
& \underline{87.96} & 88.54  & \underline{88.87} & \textcolor{Gray}{+20.6}\\

\cdashline{1-11}
\bf PretRand & $1.02\times$  & \textbf{91.27}  & \textbf{93.81} & \textbf{95.11}  &  \textbf{89.95}  & \textbf{43.54} &  \textbf{88.21} & \textbf{90.01}  & \textbf{90.23} & \textcolor{Gray}{+27.5}\\

\hline
\end{tabular}
\caption[Comparison of PretRand to baselines methods.]{\textbf{Comparison of PretRand to baselines methods}. Comparison of our method to baselines in terms of token-level accuracy for POS, CK and MST and entity-level F1 for NER (in \%) on social media test-sets. In the second column (\#params), we highlight the number of parameters of each method compared to the reference From-scratch$_{200}$ baseline. In the last column, we report the aNRG score of each method compared to the reference From-scratch$_{200}$. Best score per dataset is in bold, and the second best score is underlined.
}
\label{tab:perfs_all_data_sets}
\end{table*}

%%%%%%%%%%%%%%%%%%%%%%
%%%%%%%%%%%%%%%%%%%%%%

%-------------------------
% Table: ablation study
%-------------------------
\begin{table*}[ht!]

\footnotesize
\centering
  \begin{tabular}{| l | c c c | c | c | c c c H |}
 \hline

\multirow{2}{*}{\bf Method} & \multicolumn{3}{ c|}{\textbf{POS}} & \multicolumn{1}{c|}{\textbf{CK}} & \multicolumn{1}{c|}{\textbf{NER}} & \multicolumn{3}{ c}{\textbf{MST}} & \multirow{ 2}{*}{\textcolor{Gray}{mNRG}} \\ 
 
 & \textbf{TPoS} & \textbf{ArK} & \textbf{TweeBank}  & \textbf{TChnuk}  & \textbf{WNUT} & \textbf{Serbian} & \textbf{Slovene} & \textbf{Croatian} &\\

 \hline

\bf PretRand & \textbf{91.27} & \textbf{93.81} & \textbf{95.11}  & \textbf{89.95}  & \textbf{43.12} & \textbf{88.21} & \textbf{90.01} & \textbf{90.23}  & \textcolor{Gray}{0}\\

\hline

\bf -learnVect  &  91.11 &  93.41 &  94.71 & 89.64  & 42.76 & 88.01 & 89.83 & 90.12 &\textcolor{Gray}{-1.8}\\

\bf -learnVect -random\textsuperscript{++} & 90.84 &  93.56 &  94.26 & 89.05  & 42.70 & 87.85 & 89.39 & 89.51 &\textcolor{Gray}{-5.6} \\

\bf -learnVect -random\textsuperscript{++} -l2 norm & 90.54 &  92.19 &  93.28  & 88.66  & 41.84 & 87.66 & 88.64 & 88.49 &\textcolor{Gray}{-13.3}\\
\hline
\end{tabular}

\caption[Diagnostic analysis of the importance of each component in PretRand.]{Diagnostic analysis of the importance of each component in PretRand. Accuracy for POS, CK and MST and F1 for NER (in \%) when \textit{progressively} ablating PretRand components. }
\label{tab:perfs_pretrand_ablation}

\end{table*}

We summarise the comparison of PretRand to the above baselines in Tables \ref{tab:perfs_all_data_sets}. In the first table, we report the results of POS, CK and NER English social media datasets. In the second table, we report the results of MST on Serbian, Slovene and Croatian social media datasets. We compare the different approaches using the aNRG metric (see equation \ref{eqn:anrg}) compared to the reference From-scratch$_{200}$. First, we observe that PretRand outperforms the popular standard fine-tuning baseline significantly by +13.1 aNRG (28.8-15.7). More importantly, PretRand outperforms the challenging Ensemble method across all tasks and datasets and by +15.4 (28.8-13.4) on aNRG, while using much fewer parameters. This highlights the difference between our method and the ensemble methods. Indeed, in addition to normalisation and weighting vectors, PretRand is conceptually different since the random and pretrained branches share the WRE component. Also, the results of From-scratch$_{400}$ compared to From-scratch$_{200}$ baseline confirm that the gain brought by PretRand is not due to the supplement parameters. In the following (\cref{sec:pretrand_results_ablation_study}), we show that the gain brought by PretRand is mainly due to the shared word representation in combination with the normalisation and the learnable weighting vectors during training. Moreover, a key asset of PretRand is that it uses only 0.02\% more parameters compared to the fine-tuning baseline.

%=========================
% Ablation Study
%=========================
\subsubsection{Diagnostic Analysis of the Importance of PretRand's Components}
\label{sec:pretrand_results_ablation_study}

While in the precedent experiment we reported the best performance of PretRand, here we carry out an ablation study to diagnose the importance of each component in our proposed approach. Specifically, we successively ablate the main components of PretRand, namely, the learnable weighting vectors (learnVect), the longer training of the random branch (random++) and the normalisation (\(\ell_2\)-norm). From the results in Table \ref{tab:perfs_pretrand_ablation}, we can first observe that ablating each of them successively degrades the results across all datasets, which highlights the importance of each component. Second, the results are only marginally better than the SFT when ablating the three components from PretRand (the last line in Table \ref{tab:perfs_pretrand_ablation}). Third, ablating the normalisation layer significantly hurts the performance across all data-sets, confirming the importance of this step of making the two branches more competitive.

%%%%%%%%%%%%%%%%%%%%%%
% Table: Perfs on  all datasets With elmo
%%%%%%%%%%%%%%%%%%%%%%

\begin{table*}[ht!]

\scriptsize
\centering
  \begin{tabular}{ | l | l c c c | c c c | c |  c | c |}
 \hline

 \multirow{ 2}{*}{\textbf{ Method}} & \textbf{\multirow{2}{*}{\#}} & \textbf{\multirow{2}{*}{Char$^{\diamond\star}$}} & \textbf{\multirow{2}{*}{Word$^{\bullet\star}$}} & \textbf{\multirow{2}{*}{ELMo$^{\bullet\times}$}}  & \multicolumn{3}{ c|}{\textbf{POS (acc.)}} & \multicolumn{1}{c|}{\textbf{CK (acc.)}} & \multicolumn{1}{c}{\textbf{NER (F1.)}} & \multicolumn{1}{ |c|}{\textbf{MST (acc.)}} \\ 
 %\cline{3-11}
   & & & & & TPoS & ArK & TweeB  & TChunk & WNUT & Croatian \\
  %\cline{3-11}

 \hline
 \hline

 \multirow{7}{*}{\bf From-scratch} & \multirow{ 1}{*}{A} & \multirow{ 1}{*}{\includegraphics[scale=0.02]{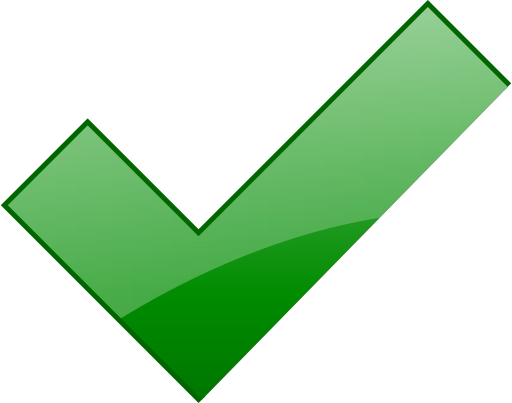}} & \multirow{ 1}{*}{\includegraphics[scale=0.03]{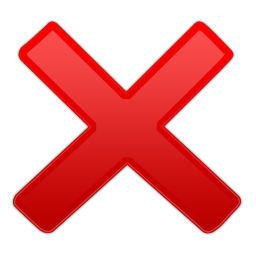}} & \multirow{1}{*}{\includegraphics[scale=0.03]{figures/no.png}} & 82.16 & 87.66 & 88.30 & 84.56 & 17.99 & 83.26 \\

\cdashline{2-11}

&\multirow{1}{*}{B} & \multirow{ 1}{*}{\includegraphics[scale=0.03]{figures/no.png}} & \multirow{ 1}{*}{\includegraphics[scale=0.02]{figures/yes.png}} & \multirow{1}{*}{\includegraphics[scale=0.03]{figures/no.png}}  & 85.21 & 88.34 & 90.63 & 84.17 & 36.58 & 80.03 \\ 

\cdashline{2-11}

&\multirow{ 1}{*}{C} & \multirow{ 1}{*}{\includegraphics[scale=0.02]{figures/yes.png}} & \multirow{1}{*}{\includegraphics[scale=0.02]{figures/yes.png}} & \multirow{1}{*}{\includegraphics[scale=0.03]{figures/no.png}}  & 86.82 & 91.10 & 91.66  & 85.96 & 36.75 & 85.67 \\

\cdashline{2-11}

&\multirow{1}{*}{D} & \multirow{ 1}{*}{\includegraphics[scale=0.03]{figures/no.png}} & \multirow{ 1}{*}{\includegraphics[scale=0.03]{figures/no.png}} & \multirow{1}{*}{\includegraphics[scale=0.02]{figures/yes.png}}  & 88.35 & 90.62 & 92.51  & 89.61 & 34.35 & 86.34 \\

\cdashline{2-11}

&\multirow{1}{*}{E} & \multirow{ 1}{*}{\includegraphics[scale=0.02]{figures/yes.png}} & \multirow{ 1}{*}{\includegraphics[scale=0.03]{figures/no.png}} & \multirow{1}{*}{\includegraphics[scale=0.02]{figures/yes.png}}   & 89.01 & 91.48 & 93.21  & 88.48 & 33.99 & 86.94 \\

\cdashline{2-11}

&\multirow{ 1}{*}{F} & \multirow{ 1}{*}{\includegraphics[scale=0.03]{figures/no.png}} & \multirow{ 1}{*}{\includegraphics[scale=0.02]{figures/yes.png}} & \multirow{1}{*}{\includegraphics[scale=0.02]{figures/yes.png}}   & 89.31 & 91.57 & 93.60  & 89.39 & 40.16 & 85.97 \\

\cdashline{2-11}

&\multirow{1}{*}{G} & \multirow{1}{*}{\includegraphics[scale=0.02]{figures/yes.png}} & \multirow{1}{*}{\includegraphics[scale=0.02]{figures/yes.png}} & \multirow{1}{*}{\includegraphics[scale=0.02]{figures/yes.png}} &  90.01 & 92.09 &  93.73  &  88.99  & 41.57  & 86.79 \\ 

\hline
\hline

 \multirow{7}{*}{\bf SFT} & \multirow{ 1}{*}{A} & \multirow{ 1}{*}{\includegraphics[scale=0.02]{figures/yes.png}} & \multirow{ 1}{*}{\includegraphics[scale=0.03]{figures/no.png}} & \multirow{1}{*}{\includegraphics[scale=0.03]{figures/no.png}} & 86.87 & 88.30 & 89.26  & 87.28 & 21.88 & 86.19 \\

\cdashline{2-11}

&\multirow{1}{*}{B} & \multirow{ 1}{*}{\includegraphics[scale=0.03]{figures/no.png}} & \multirow{ 1}{*}{\includegraphics[scale=0.02]{figures/yes.png}} & \multirow{1}{*}{\includegraphics[scale=0.03]{figures/no.png}}  & 87.61 & 89.63 & 92.31  & 87.19 & 41.50 & 83.07 \\ 

\cdashline{2-11}

&\multirow{ 1}{*}{C} & \multirow{ 1}{*}{\includegraphics[scale=0.02]{figures/yes.png}} & \multirow{1}{*}{\includegraphics[scale=0.02]{figures/yes.png}} & \multirow{1}{*}{\includegraphics[scale=0.03]{figures/no.png}}  & 89.57 & 92.09 & 93.23  & 88.86 & 41.25 & 88.79 \\

\cdashline{2-11}

&\multirow{1}{*}{D} & \multirow{ 1}{*}{\includegraphics[scale=0.03]{figures/no.png}} & \multirow{ 1}{*}{\includegraphics[scale=0.03]{figures/no.png}} & \multirow{1}{*}{\includegraphics[scale=0.02]{figures/yes.png}}  & 88.02 & 90.32 & 93.04  & 89.69 & 44.21 & 88.25 \\

\cdashline{2-11}

&\multirow{1}{*}{E} & \multirow{ 1}{*}{\includegraphics[scale=0.02]{figures/yes.png}} & \multirow{ 1}{*}{\includegraphics[scale=0.03]{figures/no.png}} & \multirow{1}{*}{\includegraphics[scale=0.02]{figures/yes.png}} & 90.18 & 91.81 & 93.53  & 90.55 & 43.98 & 88.76 \\

\cdashline{2-11}

&\multirow{ 1}{*}{F} & \multirow{ 1}{*}{\includegraphics[scale=0.03]{figures/no.png}} & \multirow{ 1}{*}{\includegraphics[scale=0.02]{figures/yes.png}} & \multirow{1}{*}{\includegraphics[scale=0.02]{figures/yes.png}}  & 88.87 & 91.83 & 93.71  & 88.82 & 45.73 & 89.28 \\

\cdashline{2-11}

&\multirow{1}{*}{G} & \multirow{1}{*}{\includegraphics[scale=0.02]{figures/yes.png}} & \multirow{1}{*}{\includegraphics[scale=0.02]{figures/yes.png}} & \multirow{1}{*}{\includegraphics[scale=0.02]{figures/yes.png}} & 90.27 & 92.73 & 94.19  & 90.75  &  46.59 & 89.00\\ 

\hline
\hline

 \multirow{7}{*}{\bf PretRand} & \multirow{ 1}{*}{A} & \multirow{ 1}{*}{\includegraphics[scale=0.02]{figures/yes.png}} & \multirow{ 1}{*}{\includegraphics[scale=0.03]{figures/no.png}} & \multirow{1}{*}{\includegraphics[scale=0.03]{figures/no.png}} & 88.01 & 90.11 & 91.16  & 88.49 & 22.12 & 87.63 \\
 
\cdashline{2-11}

&\multirow{1}{*}{B} & \multirow{ 1}{*}{\includegraphics[scale=0.03]{figures/no.png}} & \multirow{ 1}{*}{\includegraphics[scale=0.02]{figures/yes.png}} & \multirow{1}{*}{\includegraphics[scale=0.03]{figures/no.png}} & 88.56 & 90.56 & 93.99  & 88.55 & 42.87 & 93.67 \\ 
\cdashline{2-11}

&\multirow{ 1}{*}{C} & \multirow{ 1}{*}{\includegraphics[scale=0.02]{figures/yes.png}} & \multirow{1}{*}{\includegraphics[scale=0.02]{figures/yes.png}} & \multirow{1}{*}{\includegraphics[scale=0.03]{figures/no.png}}  & 91.27 & 93.81 & 95.11  & 89.95 & 43.12 & 90.23 \\

\cdashline{2-11}

&\multirow{1}{*}{D} & \multirow{ 1}{*}{\includegraphics[scale=0.03]{figures/no.png}} & \multirow{ 1}{*}{\includegraphics[scale=0.03]{figures/no.png}} & \multirow{1}{*}{\includegraphics[scale=0.02]{figures/yes.png}} & 88.15 & 90.26 & 93.41  & 89.84 & 45.54 & 88.94 \\

\cdashline{2-11}

&\multirow{1}{*}{E} & \multirow{ 1}{*}{\includegraphics[scale=0.02]{figures/yes.png}} & \multirow{ 1}{*}{\includegraphics[scale=0.03]{figures/no.png}} & \multirow{1}{*}{\includegraphics[scale=0.02]{figures/yes.png}}  & 91.12 & 92.94 & 94.89 & 91.36 & 45.13 & 89.93 \\

\cdashline{2-11}

&\multirow{ 1}{*}{F} & \multirow{ 1}{*}{\includegraphics[scale=0.03]{figures/no.png}} & \multirow{ 1}{*}{\includegraphics[scale=0.02]{figures/yes.png}} & \multirow{1}{*}{\includegraphics[scale=0.02]{figures/yes.png}} & 89.54 & 93.16 & 94.15 & 89.37 & 46.62 & 90.16 \\

\cdashline{2-11}

&\multirow{1}{*}{G} & \multirow{1}{*}{\includegraphics[scale=0.02]{figures/yes.png}} & \multirow{1}{*}{\includegraphics[scale=0.02]{figures/yes.png}} & \multirow{1}{*}{\includegraphics[scale=0.02]{figures/yes.png}} &   \textbf{91.45} & \textbf{94.18 } &  \textbf{95.22} & \textbf{91.49} & \textbf{47.33}  & \textbf{90.33} \\
\hline

\end{tabular}
\caption[Diagnosis analysis of the impact of ELMo contextual representations in training schemes: From-scratch, SFT and PretRand.]{\textbf{Diagnosis analysis of the impact of ELMo contextual representations.}. From-scratch, SFT and PretRand results, on social media test-sets, when ablating one or more type of representations. $\diamond$: from scratch, $\bullet$: pre-loaded, $\star$: trained, $\times$: frozen.}
\label{tab:ablation_study_elmo}
\end{table*}

%%%%%%%%%%%%%%%%%%%%%%
%%%%%%%%%%%%%%%%%%%%%%

%=========================
% Section: Incorporating Contextualised Words Representations
%=========================
\subsubsection{Incorporating Contextualised Word Representations}
\label{sec:results_plus_elmo}

So far in our experiments, we have used only the standard pre-loaded words embeddings and character-level embeddings in the $\mathbf{WRE}$ component. Here, we perform a further experiment that examines the effect of incorporating the ELMo contextualised word representations \citep{peters2018deep} in different tasks and training schemes (From-scratch, SFT and PretRand). Specifically, we carry out an ablation study of $\mathbf{WRE}$'s representations, namely, the standard pre-loaded words embeddings (word), character-level embeddings (char) and ELMo contextualised embeddings (ELMo). The ablation leads to 7 settings; in each, one or more representations are ablated. Results are provided in Table \ref{tab:ablation_study_elmo}, \enquote{\includegraphics[scale=0.02]{figures/yes.png}} means that the corresponding representation is used and \enquote{\includegraphics[scale=0.03]{figures/no.png}} means that it is ablated. For instance, in setting A only character-level representation is used.  

Three important observations can be highlighted. First, in training from scratch scheme, as expected, contextualised ELMo embeddings have a considerable effect on all datasets and tasks. For instance, setting D (using ELMo solely) outperforms setting C (standard concatenation between character-level and word-level embeddings), considerably on Chunking and NER and slightly on POS tagging (except ArK). Furthermore, combining ELMo embeddings to the standard concatenation between character-level and word-level embeddings (setting G) gives the best results across all tasks and social media datasets. Second, when applying our transfer learning approaches, whether SFT or PretRand, the gain brought by ELMo embeddings (setting G) compared to standard concatenation between character-level and word-level embeddings (setting C) is slight on POS tagging (in average, SFT: +0.76\% , PretRand:  +0.22\%) and Croatian MS tagging (SFT: +0.21\% , PretRand:  +0.10\%), whilst is considerable on CK (SFT: +1.89\% , PretRand:  +1.54\%) and major on NER (SFT: +5.3\% , PretRand:  +4.2\%). Finally, it should be pointed out that using ELMo slows down the training and inferences processes; it becomes 10 times slower.

%%%%%%%%%%%%%%%%%%%%%%
% Table: Comparison of PretRandto SOTA methods
%%%%%%%%%%%%%%%%%%%%%%

\begin{table*}

\footnotesize
\centering
  \begin{tabular}{l | c c c | c |  c | c c c }
 \hline
 \hline
 \multirow{ 2}{*}{\bf Method}  & \multicolumn{3}{ c|}{\textbf{POS (acc.)}} & \multicolumn{1}{c|}{\textbf{CK (acc.)}} & \multicolumn{1}{c}{\textbf{NER (F1.)}} & \multicolumn{3}{ |c}{\textbf{MST (acc.)}} \\ 
 %\cline{3-11}
  &  \bf TPoS &  \bf ArK &  \bf TweeBank &  \bf TChunk &   \bf WNUT & \bf Sr& \bf Sl & \bf Hr \\
  %\cline{3-11}

 \hline
 \hline

 CRF \citep{ritter2011named}$^\star$  & 88.3 &  n/a &  n/a & \underline{87.5} & n/a & n/a & n/a & n/a  \\
\cdashline{1-9}
 GATE \citep{derczynski2013twitter}$^\star$  & 88.69 &  n/a &  n/a & n/a & n/a & n/a & n/a & n/a  \\
\cdashline{1-9}
 GATE-bootstrap$^\star$ & 90.54 &  n/a & n/a & n/a & n/a & n/a & n/a & n/a \\
\cdashline{1-9}
 ARK tagger \citep{owoputi2013improved}$^\star$ &90.40 & 93.2 & 94.6 & n/a & n/a & n/a & n/a & n/a  \\
\cdashline{1-9}
 TPANN \citep{gui2017part}$^{\star~\times}$ & \underline{90.92} & 92.8 & n/a & n/a & n/a & n/a & n/a & n/a  \\
\cdashline{1-9}
 Flairs \citep{akbik2019pooled}$^\diamond$&  n/a &  n/a &  n/a &  n/a & 49.59 &  n/a &   n/a  &  n/a \\
\cdashline{1-9}
 MDMT \citep{mishra2019multi}$^{\diamond~\times}$ & \textbf{91.70} & 91.61 &  92.44 &  n/a &  \underline{49.86} &  n/a &   n/a  &  n/a \\
\cdashline{1-9}
 DA-LSTM \citep{gu2020data}$^\times$ & 89.16  & n/a &  n/a &  n/a & n/a &  n/a &   n/a  &  n/a \\
\cdashline{1-9}
 DA-BERT \citep{gu2020data}$^{\bullet~\times}$ & \underline{91.55}  & n/a &  n/a &  n/a & n/a &  n/a &   n/a  &  n/a \\
\cdashline{1-9}

\cdashline{1-9}
 BertTweet \citep{nguyen2020bertweet}$^{\bullet~\star}$ &  90.1 & \underline{94.1} &  \underline{95.2} &  n/a &  \textbf{54.1} &  n/a &   n/a  &  n/a \\
\cdashline{1-9}
 UH\&UC & n/a & n/a & n/a & n/a & n/a & \textbf{90.00} &  88.4  & 88.7 \\

\hline

\bf PretRand (our best)$^\diamond$  & 91.45 & \textbf{94.18 } &  \textbf{95.22} & \textbf{91.49} & 47.33  &  88.21 & \textbf{90.01}  &\textbf{90.33} \\

\hline
\hline
\end{tabular}

\caption[Comparison of PretRand to best the published SOTA methods.]{Comparison of PretRand to the best published state-of-the-art methods in terms of token-level accuracy for POS, CK and MST and F1 for NER (in \%) on social media datasets. $\diamond$: use of contextualised representations. $\bullet$: use of BERT pretrained model. $\star$: use of normalisation dictionaries, regular expressions or external knowledge. $\times$: use of a CRF classifier on top of the neural model.}
\label{tab:perfs_compare_sota}

\end{table*}

%%%%%%%%%%%%%%%%%%%%%%
%%%%%%%%%%%%%%%%%%%%%%

%=========================
% Results: Comparison to SOTA
%=========================
\subsubsection{Comparison to state-of-the-art}
\label{sec:results_compare_with_SOTA}

We compare our results to the following state-of-the-art methods:

\begin{itemize}

    \item \textbf{CRF} \citep{ritter2011named} is a Conditional Random Fields (CRF) \citep{lafferty2001conditional} based model with Brown clusters. It was jointly trained on a mixture of hand-annotated social-media texts and labelled data from the news domain, in addition to annotated IRC chat data \citep{forsythand2007lexical}.

    \item \textbf{GATE} \citep{derczynski2013twitter} is a model based on Hidden Markov Models with a set of normalisation rules, external dictionaries, lexical features and out-of-domain annotated data. The authors experimented it on TPoS, with WSJ and 32K tokens from the NPS IRC corpus. 
    They also proposed a second variety (\textbf{GATE-bootstrap}) using 1.5M additional training tokens annotated by vote-constrained bootstrapping.
    
    \item \textbf{ARK tagger} \citep{owoputi2013improved} is a model based on first-order Maximum Entropy Markov Model with greedy decoding. Brown Clusters, regular expressions and careful hand-engineered lexical features were also used. 
    
    \item \textbf{TPANN} \citep{gui2017part} is a biLSTM-CRF model that uses adversarial pre-training \citep{ganin2016domain} to leverage huge amounts of unlabelled social media texts, in addition to labelled datasets from the news domain. Next, the pretrained model is further fine-tuned on social media annotated examples. Also, regular expressions were used to tag Twitter-specific classes (hashtags, usernames, urls and @-mentions).
 
    \item \textbf{Flairs} \citep{akbik2019pooled} is a biLSTM-CRF sequence labelling architecture fed with the Pooled Contextual Embeddings \citep{akbik2018contextual} (pre-trained on character-level language models).

    \item \textbf{UH\&CU} \citep{silfverberg2018sub} is a biLSTM-based sequence labelling model for MST, jointly trained on formal and informal texts. It is similar to our base model, but used 2-stacked biLSTM layers. In addition, the particularity of UH\&CU is that the final predictions are generated as character sequences using an LSTM decoder, \textit{i.e.} a character for each morpho-syntactic feature instead of an atomic label.

    \item \textbf{Multi-dataset-multi-task (MDMT)} \citep{mishra2019multi} consists in a multi-task training of 4 NLP tasks: POS, CK, super sense tagging and NER, on 20 Tweets datasets 7 POS, 10 NER, 1 CK, and 2 super sense–tagged datasets. The model is based on a biLSTM-CRF architecture and words representations are based on the pre-trained ELMo embeddings.
    
    \item \textbf{Data Annealing (DA)} \citep{gu2020data} is a fine-tuning approach similar to our SFT baseline, but the passage from pretraining to fine-tuning is performed gradually, \textit{i.e.} the training starts with only formal text data (news) at ﬁrst; then, the proportion of the informal text data (social media) is gradually increased during the training process. They experiment with two architectural varieties, a biLSTM-based architecture (DA-LSTM) and a Transformer-based architecture (DA-BERT). In the last variety, the model is initialised with BERT$_{base}$ pretrained model (110 million parameters). %They experimented with both BERT$_{base}$ (110 million parameters) and BERT$_{large}$ (340 million parameters). 
    A CRF classifier is used as a classifier on the top of both varieties, biLSTM and BERT.

    \item \textbf{BertTweet} \citep{nguyen2020bertweet} is a large-scale model pretrained on an 80GB corpus of 850M English Tweets. The model is trained using BERT$_{base}$ \citep{devlin2019bert} architecture and following the pretraining procedure of RoBERTa \citep{liu2019roberta}. In order to perform POS tagging and NER, a randomly initialised linear prediction layer is appended on top of the last Transformer layer of BERTweet, and then the model is fine-tuned on target tasks examples. In addition, lexical dictionaries were used to normalise social media texts.

\end{itemize}

From Table \ref{tab:perfs_compare_sota}, we observe that PretRand outperforms best state-of-the-art results on POS tagging datasets (except TPoS), Chunking (+4\%), Slovene (+1.5\%) and Croatian (1.6\%) MS tagging. However, it performs worse than UH\&UC for Serbian MS tagging. This could be explained by the fact that the Serbian source dataset (news) is small compared to Slovene and Croatian, reducing the gain brought by pretraining and thus that brought by PretRand. Likewise, \citet{akbik2019pooled} outperforms our approach on NER task, in addition to using a CRF on top of the biLSTM layer, they used Contextual string embeddings that have been shown to perform better on NER than ELMo \citep{akbik2019pooled}. Also, MDMT outperforms PretRand slightly on TPoS dataset. We can observe that BERT-based approaches (DA-BERT and BertTweet) achieve strong results, especially on NER, where BertTweet begets the best state-of-the-art score. Finally, we believe that adding a CRF classification layer on top of our models will boost our results (like TPANN, MDMT, DA-LSTM and DA-BERT), as it is able to model strong dependencies between adjacent words.

%-------------------------
% Visu: Negative Transfer
%-------------------------
\begin{figure*}
\begingroup
\centering
\setlength{\tabcolsep}{2pt} % Default value: 6pt
\renewcommand{\arraystretch}{0.5} % Default value: 1
\centering
\begin{tabular}{c c}
\centering
\includegraphics[scale=0.33]{figures/negative_transfer_sft.png}
& \includegraphics[scale=0.33]{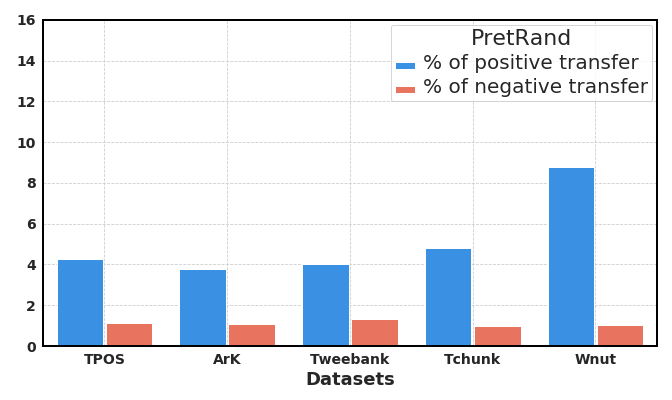}
\end{tabular}
\endgroup
\caption[Positive transfer and negative transfer brought by SFT (left) and PretRand (Right) compared to standard supervised training scheme]{Positive and negative transfers brought by SFT (left) and PretRand (Right) compared to the standard supervised training scheme (From-scratch).
}
\label{fig:negative_transfer_}
\end{figure*}
%-------------------------
%-------------------------

%-------------------------
% Visu: Sorted class-accuracy improvement
%-------------------------
\begin{figure}
\centerline{\includegraphics[scale=0.31]{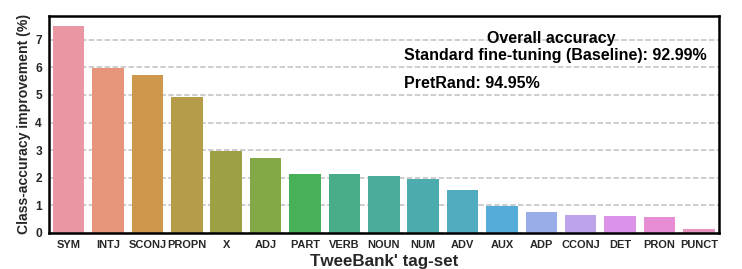}}

\caption{ Sorted class-accuracy improvement (\%) on TweeBank  of PretRand compared to fine-tuning. %\textbf{TODO: for all English datasets}
}
\label{fig:acc_class}
\end{figure}
%-------------------------

%-------------------------
% Visu: Accuracy Curve
%-------------------------
\begin{figure}
\centerline{\includegraphics[scale=0.32]{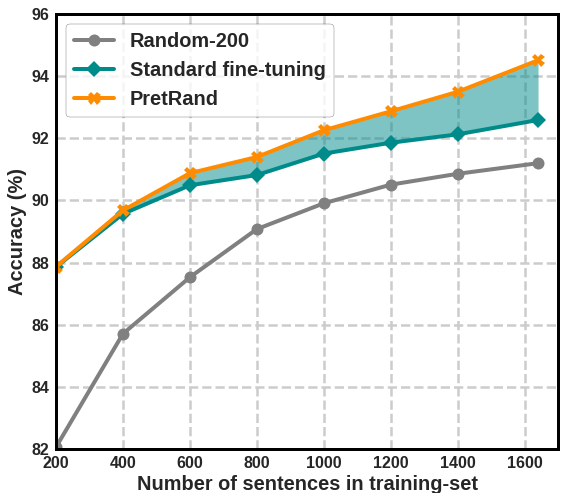}}
\caption[Performances (on dev-set of TweeBank) according different training-set sizes for the target-dataset.]{Performances (on dev-set of TweeBank) according different training-set sizes for the target-dataset. 
Transparent green highlights the difference between our PretRand and standard fine-tuning. %\textbf{TODO: for all English datasets}
}
\label{fig:accuracy_curve}
\vspace{-0.7cm}
\end{figure}
%-------------------------

%=========================
% When and where PretRand is more beneficial
%=========================
\subsubsection{When and where PretRand is most Beneficial?}
\label{sec:pretrand_when_where}

Here, we attempt to examine in which scenarios PretRand is most beneficial. We firstly explore in Figure \ref{fig:acc_class}, which class from TweeBank dataset benefits more from PretRand compared to SFT. After that, we evaluate in Figure \ref{fig:accuracy_curve} the gain on accuracy brought by PretRand compared to SFT, according to different target-datasets' sizes. We observe that PretRand has desirably a bigger gain with bigger \textit{target-task} datasets, which clearly means that the more target training-data, the more interesting our method will be. This observation may be because the random branch needs sufficient amounts of target training samples to become more competitive with the pretrained one.

%=========================
% Analysis
%=========================
\subsubsection{Negative Transfer: PretRand \textit{vs} SFT}
\label{sec:analysis_pretrand_negative_transfer}

Here, we resume the negative transfer experiment performed in section \ref{sec:proposed_method_negative_transfer_results} . Precisely, we compare the results of PretRand to those of SFT. We show in Figure \ref{fig:negative_transfer_} the results on English social media datasets, first tagged with the classic training scheme (From-scratch$_{200}$) and then \textit{using SFT} in the left plot (or \textit{using PretRand} in the right plot). Blue bars show the percentage of \textit{positive transfer}, \textit{i.e.} predictions that were wrong, but the SFT (or PretRand) changed to the correct ones, and red bars give the percentage of \textit{negative transfer}, \textit{i.e.} predictions that were tagged correctly by From-scratch$_{200}$, but using SFT (or PretRand) gives the wrong predictions. We observe the high impact of PretRand on diminishing \textit{negative transfer} vis-a-vis to SFT. Precisely, PretRand increases \textit{positive transfer} by $\sim$0.45\% and decreases the \textit{negative transfer} by $\sim$0.94\% on average.

%=========================
% Conclusion
%=========================
\section{Conclusion and Perspectives}
\label{sec:conclusion}

We have started by analysing the results of the standard fine-tuning adaptation scheme of transfer learning. First, we were interested in the hidden negative transfer that arises when transferring from the news domain to the social media domain. Indeed, negative transfer has only seldom been tackled in sequential transfer learning works in NLP. In addition, earlier research papers evoke negative transfer only when the source domain has a negative impact on the target model. We found that despite the positive gain brought by transfer learning from the high-resource news domain to the low-resource social media domain, the hidden negative transfer mitigates the final gain brought by transfer learning. Second, we carried out an interpretive analysis of the evolution, during fine-tuning, of pretrained representations. We found that while fine-tuning necessarily makes some changes during fine-tuning on social media datasets, pretrained neurons still biased by what they have learnt in the source domain. In simple words, pretrained neurons tend to conserve much information from the source domain. Some of this information is undoubtedly beneficial for the social media domain (positive transfer), but some of it is indeed harmful (negative transfer). We hypothesise that this phenomenon of biased neurons restrains the pretrained model from learning some new features specific to the target domain (social media).

Stemming from our analysis, we have introduced a novel approach,\textit{PretRand}, to overcome this problem using three main ideas: adding random units and jointly learn them with pre-trained ones; normalising the activations of both to balance their different behaviours; applying learnable weights on both predictors to let the network learn which of random or pre-trained one is better for every class. The underlying idea is to take advantage of both, target-specific features from the former and general knowledge from the latter. 
We carried out experiments on domain adaptation for 4 tasks: part-of-speech tagging, morpho-syntactic tagging, chunking and named entity recognition. Our approach exhibits performances significantly above standard fine-tuning scheme and is highly competitive when compared to the state-of-the-art.\\

\noindent
\textbf{Perspectives} 

We believe that many prosperous directions should be addressed in future research. More extensive experiments would be interesting to better understand the phenomenon of the hidden negative transfer and to confirm our observations. First, one can investigate the impact of the model's hyper-parameters (size, activation functions, learning rate, etc.) as well as regulation methods (dropout, batch normalisation, weights decay, etc.). Second, we suppose that the hidden negative transfer would be more prominent when the target dataset is too small since the pre-learned source knowledge will be more preserved. Hence, it would be interesting to assess the impact of target-training size. Third, a promising experiment would be to study the impact of the similarity between the source and the target distributions. Fourth, a fruitful direction would be to explain this hidden negative transfer using explainability methods. Notably, one can use \textit{influence functions} \citep{han2020explaining} to identify source training examples that are responsible for the negative transfer. Further, to identify text pieces of the evaluated sentence that justify a prediction with a negative transfer, one can use for instance \textit{gradients based methods} \citep{shrikumar2017learning}. 

Concerning the quantification of the change of pretrained individual neurons, it would also be interesting to perform a representation-level similarity analysis to gain more insights, as it has been shown by \citet{wu2020similarity} that representation-level similarity measures the distributional similarity while individual-level measures local similarity.

PretRand's good results on sequence labelling tasks suggest to consider other kinds of NLP tasks, e.g. sequence-to-sequence and text classification tasks. Further, as negative transfer, and thus bias, is highly arising when transferring between less-related source-target domains \citep{wang2019characterizing}, PretRand's impact would be more interesting for cross-lingual transfer. Also, in this work, we experimented PretRand adaptation scheme on models pre-trained in a supervised manner, an important step forward is to examine its scalability with other pretraining methods, \textit{e.g.} adversarial or unsupervised pretraining. In addition, the increasing omnipresence of Transformers architectures in a wide range of NLP tasks, due to their improved performances, motivates us to experiment with Transformer-based architecture instead of LSTM-based one. Last, a propitious continuity of our work to tackle the bias problem, would be to identify automatically biased neurons in the pre-trained model and proceed to a pruning of the most biased ones before fine-tuning.

\newpage
\newpage

\bibliography{eacl2021}

\newcommand{\SortNoop}[1]{}
\begin{thebibliography}{117}
\expandafter\ifx\csname natexlab\endcsname\relax\def\natexlab#1{#1}\fi

\bibitem[{Adi et~al.(2016)Adi, Kermany, Belinkov, Lavi, and
  Goldberg}]{adi2016fine}
Yossi Adi, Einat Kermany, Yonatan Belinkov, Ofer Lavi, and Yoav Goldberg. 2016.
\newblock Fine-grained analysis of sentence embeddings using auxiliary
  prediction tasks.
\newblock \emph{Proceedings of ICLR Conference Track}.

\bibitem[{Akbik et~al.(2019)Akbik, Bergmann, and Vollgraf}]{akbik2019pooled}
Alan Akbik, Tanja Bergmann, and Roland Vollgraf. 2019.
\newblock Pooled contextualized embeddings for named entity recognition.
\newblock In \emph{Proceedings of the 2019 Conference of the North American
  Chapter of the Association for Computational Linguistics: Human Language
  Technologies, Volume 1 (Long and Short Papers)}, pages 724--728.

\bibitem[{Akbik et~al.(2018)Akbik, Blythe, and Vollgraf}]{akbik2018contextual}
Alan Akbik, Duncan Blythe, and Roland Vollgraf. 2018.
\newblock Contextual string embeddings for sequence labeling.
\newblock In \emph{Proceedings of the 27th International Conference on
  Computational Linguistics}, pages 1638--1649.

\bibitem[{Bau et~al.(2019)Bau, Belinkov, Sajjad, Durrani, Dalvi, and
  Glass}]{bau2018identifying}
Anthony Bau, Yonatan Belinkov, Hassan Sajjad, Nadir Durrani, Fahim Dalvi, and
  James Glass. 2019.
\newblock Identifying and controlling important neurons in neural machine
  translation.
\newblock \emph{ICLR}.

\bibitem[{Baumann and Pierrehumbert(2014)}]{baumann2014using}
Peter Baumann and Janet~B Pierrehumbert. 2014.
\newblock Using resource-rich languages to improve morphological analysis of
  under-resourced languages.
\newblock In \emph{LREC}, pages 3355--3359.

\bibitem[{Belinkov and Glass(2019)}]{belinkov2019analysis}
Yonatan Belinkov and James Glass. 2019.
\newblock Analysis methods in neural language processing: A survey.
\newblock \emph{Transactions of the Association for Computational Linguistics},
  7:49--72.

\bibitem[{Bengio et~al.(2003)Bengio, Ducharme, Vincent, and
  Jauvin}]{bengio2003neural}
Yoshua Bengio, R{\'e}jean Ducharme, Pascal Vincent, and Christian Jauvin. 2003.
\newblock A neural probabilistic language model.
\newblock \emph{Journal of machine learning research}, 3(Feb):1137--1155.

\bibitem[{Bojanowski et~al.(2017)Bojanowski, Grave, Joulin, and
  Mikolov}]{bojanowski2017enriching}
Piotr Bojanowski, Edouard Grave, Armand Joulin, and Tomas Mikolov. 2017.
\newblock Enriching word vectors with subword information.
\newblock \emph{Transactions of the Association for Computational Linguistics},
  5:135--146.

\bibitem[{Cao et~al.(2018)Cao, Long, Wang, and Jordan}]{cao2018partial}
Zhangjie Cao, Mingsheng Long, Jianmin Wang, and Michael~I Jordan. 2018.
\newblock Partial transfer learning with selective adversarial networks.
\newblock In \emph{Proceedings of the IEEE Conference on Computer Vision and
  Pattern Recognition}, pages 2724--2732.

\bibitem[{Cer et~al.(2018)Cer, Yang, Kong, Hua, Limtiaco, John, Constant,
  Guajardo-Cespedes, Yuan, Tar et~al.}]{cer2018universal}
Daniel Cer, Yinfei Yang, Sheng-yi Kong, Nan Hua, Nicole Limtiaco, Rhomni~St
  John, Noah Constant, Mario Guajardo-Cespedes, Steve Yuan, Chris Tar, et~al.
  2018.
\newblock Universal sentence encoder for english.
\newblock In \emph{Proceedings of the 2018 Conference on Empirical Methods in
  Natural Language Processing: System Demonstrations}, pages 169--174.

\bibitem[{Che et~al.(2018)Che, Liu, Wang, Zheng, and Liu}]{che2018towards}
Wanxiang Che, Yijia Liu, Yuxuan Wang, Bo~Zheng, and Ting Liu. 2018.
\newblock Towards better ud parsing: Deep contextualized word embeddings,
  ensemble, and treebank concatenation.
\newblock In \emph{Proceedings of the CoNLL 2018 Shared Task: Multilingual
  Parsing from Raw Text to Universal Dependencies}, pages 55--64.

\bibitem[{Chen et~al.(2018)Chen, Sun, Athiwaratkun, Cardie, and
  Weinberger}]{chen2018adversarial}
Xilun Chen, Yu~Sun, Ben Athiwaratkun, Claire Cardie, and Kilian Weinberger.
  2018.
\newblock Adversarial deep averaging networks for cross-lingual sentiment
  classification.
\newblock \emph{Transactions of the Association for Computational Linguistics},
  6:557--570.

\bibitem[{Chen et~al.(2019)Chen, Wang, Fu, Long, and
  Wang}]{chen2019catastrophic}
Xinyang Chen, Sinan Wang, Bo~Fu, Mingsheng Long, and Jianmin Wang. 2019.
\newblock Catastrophic forgetting meets negative transfer: Batch spectral
  shrinkage for safe transfer learning.
\newblock In \emph{Advances in Neural Information Processing Systems}, pages
  1908--1918.

\bibitem[{Clark et~al.(2019)Clark, Khandelwal, Levy, and
  Manning}]{clark2019does}
Kevin Clark, Urvashi Khandelwal, Omer Levy, and Christopher~D Manning. 2019.
\newblock What does bert look at? an analysis of bert's attention.
\newblock \emph{arXiv preprint arXiv:1906.04341}.

\bibitem[{Coates and Ng(2011)}]{coates2011selecting}
Adam Coates and Andrew~Y Ng. 2011.
\newblock Selecting receptive fields in deep networks.
\newblock In \emph{Advances in neural information processing systems}, pages
  2528--2536.

\bibitem[{Conneau et~al.(2017)Conneau, Kiela, Schwenk, Barrault, and
  Bordes}]{conneau2017supervised}
Alexis Conneau, Douwe Kiela, Holger Schwenk, Lo{\"\i}c Barrault, and Antoine
  Bordes. 2017.
\newblock Supervised learning of universal sentence representations from
  natural language inference data.
\newblock In \emph{Proceedings of the 2017 Conference on Empirical Methods in
  Natural Language Processing}, pages 670--680.

\bibitem[{Conneau et~al.(2018)Conneau, Kruszewski, Lample, Barrault, and
  Baroni}]{conneau2018you}
Alexis Conneau, Germ{\'a}n Kruszewski, Guillaume Lample, Lo{\"\i}c Barrault,
  and Marco Baroni. 2018.
\newblock What you can cram into a single \$ \&!\#* vector: Probing sentence
  embeddings for linguistic properties.
\newblock In \emph{Proceedings of the 56th Annual Meeting of the Association
  for Computational Linguistics (Volume 1: Long Papers)}, pages 2126--2136.

\bibitem[{Derczynski et~al.(2017)Derczynski, Nichols, van Erp, and
  Limsopatham}]{derczynski2017results}
Leon Derczynski, Eric Nichols, Marieke van Erp, and Nut Limsopatham. 2017.
\newblock Results of the wnut2017 shared task on novel and emerging entity
  recognition.
\newblock In \emph{Proceedings of the 3rd Workshop on Noisy User-generated
  Text}, pages 140--147.

\bibitem[{Derczynski et~al.(2013)Derczynski, Ritter, Clark, and
  Bontcheva}]{derczynski2013twitter}
Leon Derczynski, Alan Ritter, Sam Clark, and Kalina Bontcheva. 2013.
\newblock Twitter part-of-speech tagging for all: Overcoming sparse and noisy
  data.
\newblock In \emph{Proceedings of the International Conference Recent Advances
  in Natural Language Processing RANLP 2013}, pages 198--206.

\bibitem[{Devlin et~al.(2019)Devlin, Chang, Lee, and
  Toutanova}]{devlin2019bert}
Jacob Devlin, Ming-Wei Chang, Kenton Lee, and Kristina Toutanova. 2019.
\newblock Bert: Pre-training of deep bidirectional transformers for language
  understanding.
\newblock In \emph{Proceedings of the 2019 Conference of the North American
  Chapter of the Association for Computational Linguistics: Human Language
  Technologies, Volume 1 (Long and Short Papers)}, pages 4171--4186.

\bibitem[{Dhingra et~al.(2017)Dhingra, Liu, Salakhutdinov, and
  Cohen}]{dhingra2017comparative}
Bhuwan Dhingra, Hanxiao Liu, Ruslan Salakhutdinov, and William~W Cohen. 2017.
\newblock A comparative study of word embeddings for reading comprehension.
\newblock \emph{arXiv preprint arXiv:1703.00993}.

\bibitem[{Dietterich(2000)}]{dietterich2000ensemble}
Thomas~G Dietterich. 2000.
\newblock Ensemble methods in machine learning.
\newblock In \emph{International workshop on multiple classifier systems},
  pages 1--15. Springer.

\bibitem[{Duong(2017)}]{duong2017natural}
Long Duong. 2017.
\newblock \emph{Natural language processing for resource-poor languages}.
\newblock Ph.D. thesis, University of Melbourne.

\bibitem[{Eisenstein(2019)}]{eisenstein2019measuring}
Jacob Eisenstein. 2019.
\newblock Measuring and modeling language change.
\newblock In \emph{Proceedings of the 2019 Conference of the North American
  Chapter of the Association for Computational Linguistics: Tutorials}, pages
  9--14.

\bibitem[{Fecht et~al.(2019)Fecht, Blank, and Zorn}]{fecht2019sequential}
Pascal Fecht, Sebastian Blank, and Hans-Peter Zorn. 2019.
\newblock Sequential transfer learning in nlp for german text summarization.

\bibitem[{Forsythand and Martell(2007)}]{forsythand2007lexical}
Eric~N Forsythand and Craig~H Martell. 2007.
\newblock Lexical and discourse analysis of online chat dialog.
\newblock In \emph{Semantic Computing, 2007. ICSC 2007. International
  Conference on}, pages 19--26. IEEE.

\bibitem[{Ganin et~al.(2016)Ganin, Ustinova, Ajakan, Germain, Larochelle,
  Laviolette, Marchand, and Lempitsky}]{ganin2016domain}
Yaroslav Ganin, Evgeniya Ustinova, Hana Ajakan, Pascal Germain, Hugo
  Larochelle, Fran{\c{c}}ois Laviolette, Mario Marchand, and Victor Lempitsky.
  2016.
\newblock Domain-adversarial training of neural networks.
\newblock \emph{The Journal of Machine Learning Research}, 17(1):2096--2030.

\bibitem[{Ge et~al.(2014)Ge, Gao, Ngo, Li, and Zhang}]{ge2014handling}
Liang Ge, Jing Gao, Hung Ngo, Kang Li, and Aidong Zhang. 2014.
\newblock On handling negative transfer and imbalanced distributions in
  multiple source transfer learning.
\newblock \emph{Statistical Analysis and Data Mining: The ASA Data Science
  Journal}, 7(4):254--271.

\bibitem[{Giorgi and Bader(2018)}]{giorgi2018transfer}
John~M Giorgi and Gary~D Bader. 2018.
\newblock Transfer learning for biomedical named entity recognition with neural
  networks.
\newblock \emph{Bioinformatics}, 34(23):4087--4094.

\bibitem[{Girshick et~al.(2014)Girshick, Donahue, Darrell, and
  Malik}]{girshick2014rich}
Ross Girshick, Jeff Donahue, Trevor Darrell, and Jitendra Malik. 2014.
\newblock Rich feature hierarchies for accurate object detection and semantic
  segmentation.
\newblock In \emph{Proceedings of the IEEE conference on computer vision and
  pattern recognition}, pages 580--587.

\bibitem[{Graves et~al.(2013)Graves, Jaitly, and Mohamed}]{graves2013hybrid}
Alex Graves, Navdeep Jaitly, and Abdel-rahman Mohamed. 2013.
\newblock Hybrid speech recognition with deep bidirectional lstm.
\newblock In \emph{2013 IEEE workshop on automatic speech recognition and
  understanding}, pages 273--278. IEEE.

\bibitem[{Gu and Yu(2020)}]{gu2020data}
Jing Gu and Zhou Yu. 2020.
\newblock Data annealing for informal language understanding tasks.
\newblock \emph{EMNLP2020 Findings}.

\bibitem[{Gui et~al.(2018)Gui, Xu, Lu, Du, and Zhou}]{gui2018negative}
Lin Gui, Ruifeng Xu, Qin Lu, Jiachen Du, and Yu~Zhou. 2018.
\newblock Negative transfer detection in transductive transfer learning.
\newblock \emph{International Journal of Machine Learning and Cybernetics},
  9(2):185--197.

\bibitem[{Gui et~al.(2017)Gui, Zhang, Huang, Peng, and Huang}]{gui2017part}
Tao Gui, Qi~Zhang, Haoran Huang, Minlong Peng, and Xuan-Jing Huang. 2017.
\newblock Part-of-speech tagging for twitter with adversarial neural networks.
\newblock In \emph{Proceedings of the 2017 Conference on Empirical Methods in
  Natural Language Processing}, pages 2411--2420.

\bibitem[{Han et~al.(2020)Han, Wallace, and Tsvetkov}]{han2020explaining}
Xiaochuang Han, Byron~C Wallace, and Yulia Tsvetkov. 2020.
\newblock Explaining black box predictions and unveiling data artifacts through
  influence functions.
\newblock \emph{arXiv preprint arXiv:2005.06676}.

\bibitem[{Horsmann(2018)}]{horsmann2018robust}
Tobias Horsmann. 2018.
\newblock \emph{Robust part-of-speech tagging of social media text}.
\newblock Ph.D. thesis.

\bibitem[{Hotelling(1992)}]{hotelling1992relations}
Harold Hotelling. 1992.
\newblock Relations between two sets of variates.
\newblock In \emph{Breakthroughs in statistics}, pages 162--190. Springer.

\bibitem[{Houlsby et~al.(2019)Houlsby, Giurgiu, Jastrzebski, Morrone,
  De~Laroussilhe, Gesmundo, Attariyan, and Gelly}]{houlsby2019parameter}
Neil Houlsby, Andrei Giurgiu, Stanislaw Jastrzebski, Bruna Morrone, Quentin
  De~Laroussilhe, Andrea Gesmundo, Mona Attariyan, and Sylvain Gelly. 2019.
\newblock Parameter-efficient transfer learning for nlp.
\newblock In \emph{International Conference on Machine Learning}, pages
  2790--2799.

\bibitem[{Hubel and Wiesel(1965)}]{hubel1965receptive}
David~H Hubel and Torsten~N Wiesel. 1965.
\newblock Receptive fields and functional architecture in two nonstriate visual
  areas (18 and 19) of the cat.
\newblock \emph{Journal of neurophysiology}, 28(2):229--289.

\bibitem[{K{\'a}d{\'a}r et~al.(2017)K{\'a}d{\'a}r, Chrupa{\l}a, and
  Alishahi}]{kadar2017representation}
Akos K{\'a}d{\'a}r, Grzegorz Chrupa{\l}a, and Afra Alishahi. 2017.
\newblock Representation of linguistic form and function in recurrent neural
  networks.
\newblock \emph{Computational Linguistics}, 43(4):761--780.

\bibitem[{Karpathy et~al.(2016)Karpathy, Johnson, and
  Fei-Fei}]{karpathy2016visualizing}
Andrej Karpathy, Justin Johnson, and Li~Fei-Fei. 2016.
\newblock Visualizing and understanding recurrent networks.
\newblock \emph{Proceedings of ICLR Conference Track}.

\bibitem[{Kocmi(2020)}]{kocmi2020exploring}
Tom Kocmi. 2020.
\newblock \emph{Exploring Benefits of Transfer Learning in Neural Machine
  Translation}.
\newblock Ph.D. thesis, Univerzita Karlova, Matematicko-fyzik{\'a}ln{\'\i}
  fakulta.

\bibitem[{Lafferty et~al.(2001)Lafferty, McCallum, and
  Pereira}]{lafferty2001conditional}
John~D Lafferty, Andrew McCallum, and Fernando~CN Pereira. 2001.
\newblock Conditional random fields: Probabilistic models for segmenting and
  labeling sequence data.
\newblock In \emph{Proceedings of the Eighteenth International Conference on
  Machine Learning}, pages 282--289.

\bibitem[{Lakretz et~al.(2019)Lakretz, Kruszewski, Desbordes, Hupkes, Dehaene,
  and Baroni}]{lakretz2019emergence}
Yair Lakretz, Germ{\'a}n Kruszewski, Th{\'e}o Desbordes, Dieuwke Hupkes,
  Stanislas Dehaene, and Marco Baroni. 2019.
\newblock The emergence of number and syntax units in lstm language models.
\newblock In \emph{Proceedings of the 2019 Conference of the North American
  Chapter of the Association for Computational Linguistics: Human Language
  Technologies, Volume 1 (Long and Short Papers)}, pages 11--20.

\bibitem[{Lample et~al.(2018)Lample, Conneau, Ranzato, Denoyer, and
  J{\'e}gou}]{lample2018word}
Guillaume Lample, Alexis Conneau, Marc'Aurelio Ranzato, Ludovic Denoyer, and
  Herv{\'e} J{\'e}gou. 2018.
\newblock Word translation without parallel data.
\newblock \emph{ICLR2018}.

\bibitem[{Li et~al.(2015)Li, Yosinski, Clune, Lipson, and
  Hopcroft}]{li2015convergent}
Yixuan Li, Jason Yosinski, Jeff Clune, Hod Lipson, and John Hopcroft. 2015.
\newblock Convergent learning: Do different neural networks learn the same
  representations?
\newblock In \emph{Feature Extraction: Modern Questions and Challenges}, pages
  196--212.

\bibitem[{Lin and Lu(2018)}]{lin2018neural}
Bill~Yuchen Lin and Wei Lu. 2018.
\newblock Neural adaptation layers for cross-domain named entity recognition.
\newblock In \emph{Proceedings of the 2018 Conference on Empirical Methods in
  Natural Language Processing}, pages 2012--2022.

\bibitem[{Liu et~al.(2019{\natexlab{a}})Liu, Gardner, Belinkov, Peters, and
  Smith}]{liu2019linguistic}
Nelson~F Liu, Matt Gardner, Yonatan Belinkov, Matthew~E Peters, and Noah~A
  Smith. 2019{\natexlab{a}}.
\newblock Linguistic knowledge and transferability of contextual
  representations.
\newblock In \emph{Proceedings of the 2019 Conference of the North American
  Chapter of the Association for Computational Linguistics: Human Language
  Technologies, Volume 1 (Long and Short Papers)}, pages 1073--1094.

\bibitem[{Liu et~al.(2015)Liu, Rabinovich, and Berg}]{liu2015parsenet}
Wei Liu, Andrew Rabinovich, and Alexander~C Berg. 2015.
\newblock Parsenet: Looking wider to see better.
\newblock \emph{arXiv preprint arXiv:1506.04579}.

\bibitem[{Liu et~al.(2018)Liu, Zhu, Che, Qin, Schneider, and
  Smith}]{liu2018parsing}
Yijia Liu, Yi~Zhu, Wanxiang Che, Bing Qin, Nathan Schneider, and Noah~A Smith.
  2018.
\newblock Parsing tweets into universal dependencies.
\newblock In \emph{Proceedings of the 2018 Conference of the North American
  Chapter of the Association for Computational Linguistics: Human Language
  Technologies, Volume 1 (Long Papers)}, pages 965--975.

\bibitem[{Liu et~al.(2019{\natexlab{b}})Liu, Ott, Goyal, Du, Joshi, Chen, Levy,
  Lewis, Zettlemoyer, and Stoyanov}]{liu2019roberta}
Yinhan Liu, Myle Ott, Naman Goyal, Jingfei Du, Mandar Joshi, Danqi Chen, Omer
  Levy, Mike Lewis, Luke Zettlemoyer, and Veselin Stoyanov. 2019{\natexlab{b}}.
\newblock Roberta: A robustly optimized bert pretraining approach.
\newblock \emph{arXiv preprint arXiv:1907.11692}.

\bibitem[{Ma and Hovy(2016)}]{ma2016end}
Xuezhe Ma and Eduard Hovy. 2016.
\newblock End-to-end sequence labeling via bi-directional lstm-cnns-crf.
\newblock In \emph{Proceedings of the 54th Annual Meeting of the Association
  for Computational Linguistics (Volume 1: Long Papers)}, pages 1064--1074.

\bibitem[{Marcus et~al.(1993)Marcus, Santorini, and
  Marcinkiewicz}]{marcus1993building}
Mitchell Marcus, Beatrice Santorini, and Mary~Ann Marcinkiewicz. 1993.
\newblock Building a large annotated corpus of english: The penn treebank.
\newblock Technical report, University of Pennsylvania Department of Computer
  and Information Science.

\bibitem[{M{\"a}rz et~al.(2019)M{\"a}rz, Trautmann, and Roth}]{marz2019domain}
Luisa M{\"a}rz, Dietrich Trautmann, and Benjamin Roth. 2019.
\newblock Domain adaptation for part-of-speech tagging of noisy user-generated
  text.
\newblock In \emph{Proceedings of the 2019 Conference of the North American
  Chapter of the Association for Computational Linguistics: Human Language
  Technologies, Volume 1 (Long and Short Papers)}, pages 3415--3420.

\bibitem[{McCann et~al.(2017)McCann, Bradbury, Xiong, and
  Socher}]{mccann2017learned}
Bryan McCann, James Bradbury, Caiming Xiong, and Richard Socher. 2017.
\newblock Learned in translation: Contextualized word vectors.
\newblock In \emph{Advances in Neural Information Processing Systems}, pages
  6294--6305.

\bibitem[{Meftah et~al.(2018{\natexlab{a}})Meftah, Semmar, and
  Sadat}]{meftah2018neural}
Sara Meftah, Nasredine Semmar, and Fatiha Sadat. 2018{\natexlab{a}}.
\newblock A neural network model for part-of-speech tagging of social media
  texts.
\newblock In \emph{Proceedings of the Eleventh International Conference on
  Language Resources and Evaluation (LREC 2018)}.

\bibitem[{Meftah et~al.(2018{\natexlab{b}})Meftah, Semmar, Sadat, and
  Raaijmakers}]{meftah2018using}
Sara Meftah, Nasredine Semmar, Fatiha Sadat, and Stephan Raaijmakers.
  2018{\natexlab{b}}.
\newblock Using neural transfer learning for morpho-syntactic tagging of
  south-slavic languages tweets.
\newblock In \emph{Proceedings of the Fifth Workshop on NLP for Similar
  Languages, Varieties and Dialects (VarDial 2018)}, pages 235--243.

\bibitem[{Meftah et~al.(2020)Meftah, Semmar, Tahiri, Tamaazousti, Essafi, and
  Sadat}]{meftah2020multi}
Sara Meftah, Nasredine Semmar, Mohamed-Ayoub Tahiri, Youssef Tamaazousti,
  Hassane Essafi, and Fatiha Sadat. 2020.
\newblock Multi-task supervised pretraining for neural domain adaptation.
\newblock In \emph{Proceedings of the Eighth International Workshop on Natural
  Language Processing for Social Media}, pages 61--71.

\bibitem[{Meftah et~al.(2017)Meftah, Semmar, Zennaki, and
  Sadat}]{meftah2017supervised}
Sara Meftah, Nasredine Semmar, Othmane Zennaki, and Fatiha Sadat. 2017.
\newblock Supervised transfer learning for sequence tagging of
  user-generated-content in social media.
\newblock In \emph{Language and Technology Conference}, pages 43--57. Springer.

\bibitem[{Meftah et~al.(2019)Meftah, Tamaazousti, Semmar, Essafi, and
  Sadat}]{meftah2019joint}
Sara Meftah, Youssef Tamaazousti, Nasredine Semmar, Hassane Essafi, and Fatiha
  Sadat. 2019.
\newblock Joint learning of pre-trained and random units for domain adaptation
  in part-of-speech tagging.
\newblock In \emph{Proceedings of the 2019 Conference of the North American
  Chapter of the Association for Computational Linguistics: Human Language
  Technologies, Volume 1 (Long and Short Papers)}, pages 4107--4112.

\bibitem[{Merchant et~al.(2020)Merchant, Rahimtoroghi, Pavlick, and
  Tenney}]{merchant2020hapepns}
Amil Merchant, Elahe Rahimtoroghi, Ellie Pavlick, and Ian Tenney. 2020.
\newblock What happens to bert embeddings during fine-tuning?
\newblock \emph{arXiv preprint arXiv:2004.14448}.

\bibitem[{Mikolov et~al.(2013)Mikolov, Chen, Corrado, and
  Dean}]{mikolov2013efficient}
Tomas Mikolov, Kai Chen, Greg Corrado, and Jeffrey Dean. 2013.
\newblock Efficient estimation of word representations in vector space.
\newblock \emph{Proceedings of the International Conference on Learning
  Representations (ICLR 2013)}.

\bibitem[{Mishra(2019)}]{mishra2019multi}
Shubhanshu Mishra. 2019.
\newblock Multi-dataset-multi-task neural sequence tagging for information
  extraction from tweets.
\newblock In \emph{Proceedings of the 30th ACM Conference on Hypertext and
  Social Media}, pages 283--284. ACM.

\bibitem[{Morcos et~al.(2018)Morcos, Raghu, and Bengio}]{morcos2018insights}
Ari Morcos, Maithra Raghu, and Samy Bengio. 2018.
\newblock Insights on representational similarity in neural networks with
  canonical correlation.
\newblock In \emph{Advances in Neural Information Processing Systems}, pages
  5727--5736.

\bibitem[{Mou et~al.(2016)Mou, Meng, Yan, Li, Xu, Zhang, and
  Jin}]{mou2016transferable}
Lili Mou, Zhao Meng, Rui Yan, Ge~Li, Yan Xu, Lu~Zhang, and Zhi Jin. 2016.
\newblock How transferable are neural networks in nlp applications?
\newblock In \emph{Proceedings of the 2016 Conference on Empirical Methods in
  Natural Language Processing}, pages 479--489.

\bibitem[{Nguyen et~al.(2020)Nguyen, Vu, and Nguyen}]{nguyen2020bertweet}
Dat~Quoc Nguyen, Thanh Vu, and Anh~Tuan Nguyen. 2020.
\newblock Bertweet: A pre-trained language model for english tweets.
\newblock \emph{arXiv preprint arXiv:2005.10200}.

\bibitem[{Nie et~al.(2017)Nie, Bennett, and Goodman}]{nie2017dissent}
Allen Nie, Erin~D Bennett, and Noah~D Goodman. 2017.
\newblock Dissent: Sentence representation learning from explicit discourse
  relations.
\newblock \emph{arXiv preprint arXiv:1710.04334}.

\bibitem[{O'Neill(2019)}]{o2019learning}
James O'Neill. 2019.
\newblock Learning to avoid negative transfer in few shot transfer learning.
\newblock \emph{openreview.net}.

\bibitem[{Owoputi et~al.(2013)Owoputi, O’Connor, Dyer, Gimpel, Schneider, and
  Smith}]{owoputi2013improved}
Olutobi Owoputi, Brendan O’Connor, Chris Dyer, Kevin Gimpel, Nathan
  Schneider, and Noah~A Smith. 2013.
\newblock Improved part-of-speech tagging for online conversational text with
  word clusters.
\newblock In \emph{Proceedings of the 2013 conference of the North American
  chapter of the association for computational linguistics: human language
  technologies}, pages 380--390.

\bibitem[{Pan et~al.(2010)Pan, Yang et~al.}]{pan2010survey}
Sinno~Jialin Pan, Qiang Yang, et~al. 2010.
\newblock A survey on transfer learning.
\newblock \emph{IEEE Transactions on knowledge and data engineering},
  22(10):1345--1359.

\bibitem[{Paszke et~al.(2017)Paszke, Gross, Chintala, Chanan, Yang, DeVito,
  Lin, Desmaison, Antiga, and Lerer}]{paszke2017automatic}
Adam Paszke, Sam Gross, Soumith Chintala, Gregory Chanan, Edward Yang, Zachary
  DeVito, Zeming Lin, Alban Desmaison, Luca Antiga, and Adam Lerer. 2017.
\newblock Automatic differentiation in pytorch.

\bibitem[{Pennington et~al.(2014)Pennington, Socher, and
  Manning}]{pennington2014glove}
Jeffrey Pennington, Richard Socher, and Christopher Manning. 2014.
\newblock Glove: Global vectors for word representation.
\newblock In \emph{Proceedings of the 2014 conference on empirical methods in
  natural language processing (EMNLP)}, pages 1532--1543.

\bibitem[{Peters et~al.(2017)Peters, Ammar, Bhagavatula, and
  Power}]{peters2017semi}
Matthew Peters, Waleed Ammar, Chandra Bhagavatula, and Russell Power. 2017.
\newblock Semi-supervised sequence tagging with bidirectional language models.
\newblock In \emph{Proceedings of the 55th Annual Meeting of the Association
  for Computational Linguistics (Volume 1: Long Papers)}, pages 1756--1765.

\bibitem[{Peters et~al.(2019)Peters, Ruder, and Smith}]{peters2019tune}
Matthew Peters, Sebastian Ruder, and Noah~A Smith. 2019.
\newblock To tune or not to tune? adapting pretrained representations to
  diverse tasks.
\newblock \emph{arXiv preprint arXiv:1903.05987}.

\bibitem[{Peters et~al.(2018)Peters, Neumann, Iyyer, Gardner, Clark, Lee, and
  Zettlemoyer}]{peters2018deep}
Matthew~E Peters, Mark Neumann, Mohit Iyyer, Matt Gardner, Christopher Clark,
  Kenton Lee, and Luke Zettlemoyer. 2018.
\newblock Deep contextualized word representations.
\newblock In \emph{Proceedings of NAACL-HLT}, pages 2227--2237.

\bibitem[{Pfeiffer et~al.(2020{\natexlab{a}})Pfeiffer, Kamath, R{\"u}ckl{\'e},
  Cho, and Gurevych}]{pfeiffer2020adapterfusion}
Jonas Pfeiffer, Aishwarya Kamath, Andreas R{\"u}ckl{\'e}, Kyunghyun Cho, and
  Iryna Gurevych. 2020{\natexlab{a}}.
\newblock Adapterfusion: Non-destructive task composition for transfer
  learning.
\newblock \emph{arXiv preprint arXiv:2005.00247}.

\bibitem[{Pfeiffer et~al.(2020{\natexlab{b}})Pfeiffer, R{\"u}ckl{\'e}, Poth,
  Kamath, Vuli{\'c}, Ruder, Cho, and Gurevych}]{pfeiffer2020adapterhub}
Jonas Pfeiffer, Andreas R{\"u}ckl{\'e}, Clifton Poth, Aishwarya Kamath, Ivan
  Vuli{\'c}, Sebastian Ruder, Kyunghyun Cho, and Iryna Gurevych.
  2020{\natexlab{b}}.
\newblock Adapterhub: A framework for adapting transformers.
\newblock \emph{arXiv preprint arXiv:2007.07779}.

\bibitem[{Pfeiffer et~al.(2020{\natexlab{c}})Pfeiffer, Vuli{\'c}, Gurevych, and
  Ruder}]{pfeiffer2020mad}
Jonas Pfeiffer, Ivan Vuli{\'c}, Iryna Gurevych, and Sebastian Ruder.
  2020{\natexlab{c}}.
\newblock Mad-x: An adapter-based framework for multi-task cross-lingual
  transfer.
\newblock \emph{arXiv preprint arXiv:2005.00052}.

\bibitem[{Plank et~al.(2016)Plank, S{\o}gaard, and
  Goldberg}]{plank-etal-2016-multilingual}
Barbara Plank, Anders S{\o}gaard, and Yoav Goldberg. 2016.
\newblock \href {https://doi.org/10.18653/v1/P16-2067} {Multilingual
  part-of-speech tagging with bidirectional long short-term memory models and
  auxiliary loss}.
\newblock In \emph{Proceedings of the 54th Annual Meeting of the Association
  for Computational Linguistics (Volume 2: Short Papers)}, pages 412--418,
  Berlin, Germany. Association for Computational Linguistics.

\bibitem[{Radford et~al.(2017)Radford, Jozefowicz, and
  Sutskever}]{radford2017learning}
Alec Radford, Rafal Jozefowicz, and Ilya Sutskever. 2017.
\newblock Learning to generate reviews and discovering sentiment.
\newblock \emph{arXiv preprint arXiv:1704.01444}.

\bibitem[{Raffel et~al.(2019)Raffel, Shazeer, Roberts, Lee, Narang, Matena,
  Zhou, Li, and Liu}]{2019t5}
Colin Raffel, Noam Shazeer, Adam Roberts, Katherine Lee, Sharan Narang, Michael
  Matena, Yanqi Zhou, Wei Li, and Peter~J. Liu. 2019.
\newblock \href {http://arxiv.org/abs/1910.10683} {Exploring the limits of
  transfer learning with a unified text-to-text transformer}.
\newblock \emph{arXiv e-prints}.

\bibitem[{Raghu et~al.(2017)Raghu, Gilmer, Yosinski, and
  Sohl-Dickstein}]{raghu2017svcca}
Maithra Raghu, Justin Gilmer, Jason Yosinski, and Jascha Sohl-Dickstein. 2017.
\newblock Svcca: Singular vector canonical correlation analysis for deep
  learning dynamics and interpretability.
\newblock In \emph{Advances in Neural Information Processing Systems}, pages
  6076--6085.

\bibitem[{Raghu et~al.(2019)Raghu, Zhang, Kleinberg, and
  Bengio}]{raghu2019transfusion}
Maithra Raghu, Chiyuan Zhang, Jon Kleinberg, and Samy Bengio. 2019.
\newblock Transfusion: Understanding transfer learning with applications to
  medical imaging.
\newblock \emph{NeurIPS}.

\bibitem[{Ramachandran et~al.(2017)Ramachandran, Liu, and
  Le}]{ramachandran2017unsupervised}
Prajit Ramachandran, Peter~J Liu, and Quoc Le. 2017.
\newblock Unsupervised pretraining for sequence to sequence learning.
\newblock In \emph{Proceedings of the 2017 Conference on Empirical Methods in
  Natural Language Processing}, pages 383--391.

\bibitem[{Rebuffi et~al.(2017)Rebuffi, Bilen, and
  Vedaldi}]{rebuffi2017learning}
Sylvestre-Alvise Rebuffi, Hakan Bilen, and Andrea Vedaldi. 2017.
\newblock Learning multiple visual domains with residual adapters.
\newblock In \emph{Advances in Neural Information Processing Systems}, pages
  506--516.

\bibitem[{Ritter et~al.(2011)Ritter, Clark, Etzioni et~al.}]{ritter2011named}
Alan Ritter, Sam Clark, Oren Etzioni, et~al. 2011.
\newblock Named entity recognition in tweets: an experimental study.
\newblock In \emph{Proceedings of the conference on empirical methods in
  natural language processing}, pages 1524--1534. Association for Computational
  Linguistics.

\bibitem[{Rosenstein et~al.(2005)Rosenstein, Marx, Kaelbling, and
  Dietterich}]{rosenstein2005transfer}
Michael~T Rosenstein, Zvika Marx, Leslie~Pack Kaelbling, and Thomas~G
  Dietterich. 2005.
\newblock To transfer or not to transfer.
\newblock In \emph{In NIPS’05 Workshop, Inductive Transfer: 10 Years Later}.
  Citeseer.

\bibitem[{Ruder(2019)}]{ruder2019neural}
Sebastian Ruder. 2019.
\newblock \emph{Neural Transfer Learning for Natural Language Processing}.
\newblock Ph.D. thesis, NATIONAL UNIVERSITY OF IRELAND, GALWAY.

\bibitem[{Saphra and Lopez(2019)}]{saphra2019understanding}
Naomi Saphra and Adam Lopez. 2019.
\newblock Understanding learning dynamics of language models with svcca.
\newblock In \emph{Proceedings of the 2019 Conference of the North American
  Chapter of the Association for Computational Linguistics: Human Language
  Technologies, Volume 1 (Long and Short Papers)}, pages 3257--3267.

\bibitem[{Schumacher and Dredze(2019)}]{schumacher2019learning}
Elliot Schumacher and Mark Dredze. 2019.
\newblock Learning unsupervised contextual representations for medical synonym
  discovery.
\newblock \emph{JAMIA Open}.

\bibitem[{Seah et~al.(2012)Seah, Ong, and Tsang}]{seah2012combating}
Chun-Wei Seah, Yew-Soon Ong, and Ivor~W Tsang. 2012.
\newblock Combating negative transfer from predictive distribution differences.
\newblock \emph{IEEE transactions on cybernetics}, 43(4):1153--1165.

\bibitem[{Shi et~al.(2016)Shi, Padhi, and Knight}]{shi2016does}
Xing Shi, Inkit Padhi, and Kevin Knight. 2016.
\newblock Does string-based neural mt learn source syntax?
\newblock In \emph{Proceedings of the 2016 Conference on Empirical Methods in
  Natural Language Processing}, pages 1526--1534.

\bibitem[{Shrikumar et~al.(2017)Shrikumar, Greenside, and
  Kundaje}]{shrikumar2017learning}
Avanti Shrikumar, Peyton Greenside, and Anshul Kundaje. 2017.
\newblock Learning important features through propagating activation
  differences.
\newblock In \emph{International Conference on Machine Learning}, pages
  3145--3153.

\bibitem[{Silfverberg and Drobac(2018)}]{silfverberg2018sub}
Miikka Silfverberg and Senka Drobac. 2018.
\newblock Sub-label dependencies for neural morphological tagging--the joint
  submission of university of colorado and university of helsinki for vardial
  2018.
\newblock In \emph{Proceedings of the Fifth Workshop on NLP for Similar
  Languages, Varieties and Dialects (VarDial 2018)}, pages 37--45.

\bibitem[{Subramanian et~al.(2018)Subramanian, Trischler, Bengio, and
  Pal}]{subramanian2018learning}
Sandeep Subramanian, Adam Trischler, Yoshua Bengio, and Christopher~J Pal.
  2018.
\newblock Learning general purpose distributed sentence representations via
  large scale multi-task learning.
\newblock \emph{arXiv preprint arXiv:1804.00079}.

\bibitem[{Tamaazousti(2018)}]{tamaazousti2018universality}
Youssef Tamaazousti. 2018.
\newblock On the universality of visual and multimodal representations.
\newblock \emph{PhD thesis}.

\bibitem[{Tamaazousti et~al.(2017)Tamaazousti, Le~Borgne, and
  Hudelot}]{tamaazousti2017mucale}
Youssef Tamaazousti, Herv{\'e} Le~Borgne, and C{\'e}line Hudelot. 2017.
\newblock Mucale-net: Multi categorical-level networks to generate more
  discriminating features.
\newblock In \emph{IEEE Computer Vision and Pattern Recognition}.

\bibitem[{Tamaazousti et~al.(2019)Tamaazousti, Le~Borgne, Hudelot, Seddik, and
  Tamaazousti}]{tamaazousti2018universal}
Youssef Tamaazousti, Herv\'e Le~Borgne, C\'eline Hudelot, Mohamed El~Amine
  Seddik, and Mohamed Tamaazousti. 2019.
\newblock Learning more universal representations for transfer-learning.
\newblock \emph{IEEE Transactions on Pattern Analysis and Machine
  Intelligence}.

\bibitem[{Tjong Kim~Sang and Buchholz(2000)}]{tjong2000introduction}
Erik~F Tjong Kim~Sang and Sabine Buchholz. 2000.
\newblock Introduction to the conll-2000 shared task: chunking.
\newblock In \emph{Proceedings of the 2nd workshop on Learning language in
  logic and the 4th conference on Computational natural language
  learning-Volume 7}, pages 127--132.

\bibitem[{Tjong Kim~Sang and De~Meulder(2003)}]{tjong2003introduction}
Erik~F Tjong Kim~Sang and Fien De~Meulder. 2003.
\newblock Introduction to the conll-2003 shared task: language-independent
  named entity recognition.
\newblock In \emph{Proceedings of the seventh conference on Natural language
  learning at HLT-NAACL 2003-Volume 4}, pages 142--147.

\bibitem[{Torrey and Shavlik(2010)}]{torrey2010transfer}
Lisa Torrey and Jude Shavlik. 2010.
\newblock Transfer learning.
\newblock In \emph{Handbook of research on machine learning applications and
  trends: algorithms, methods, and techniques}, pages 242--264. IGI global.

\bibitem[{Uurtio et~al.(2018)Uurtio, Monteiro, Kandola, Shawe-Taylor,
  Fernandez-Reyes, and Rousu}]{uurtio2018tutorial}
Viivi Uurtio, Jo{\~a}o~M Monteiro, Jaz Kandola, John Shawe-Taylor, Delmiro
  Fernandez-Reyes, and Juho Rousu. 2018.
\newblock A tutorial on canonical correlation methods.
\newblock \emph{ACM Computing Surveys (CSUR)}, 50(6):95.

\bibitem[{Vaswani et~al.(2017)Vaswani, Shazeer, Parmar, Uszkoreit, Jones,
  Gomez, Kaiser, and Polosukhin}]{vaswani2017attention}
Ashish Vaswani, Noam Shazeer, Niki Parmar, Jakob Uszkoreit, Llion Jones,
  Aidan~N Gomez, {\L}ukasz Kaiser, and Illia Polosukhin. 2017.
\newblock Attention is all you need.
\newblock In \emph{Advances in neural information processing systems}, pages
  5998--6008.

\bibitem[{Wang et~al.(2017)Wang, Ramanan, and Hebert}]{wang2017growing}
Yu-Xiong Wang, Deva Ramanan, and Martial Hebert. 2017.
\newblock Growing a brain: Fine-tuning by increasing model capacity.
\newblock In \emph{CVPR}, pages 2471--2480.

\bibitem[{Wang et~al.(2019)Wang, Dai, P{\'o}czos, and
  Carbonell}]{wang2019characterizing}
Zirui Wang, Zihang Dai, Barnab{\'a}s P{\'o}czos, and Jaime Carbonell. 2019.
\newblock Characterizing and avoiding negative transfer.
\newblock In \emph{Proceedings of the IEEE Conference on Computer Vision and
  Pattern Recognition}, pages 11293--11302.

\bibitem[{Wiese et~al.(2017)Wiese, Weissenborn, and Neves}]{wiese2017neural}
Georg Wiese, Dirk Weissenborn, and Mariana Neves. 2017.
\newblock Neural domain adaptation for biomedical question answering.
\newblock In \emph{Proceedings of the 21st Conference on Computational Natural
  Language Learning (CoNLL 2017)}, pages 281--289.

\bibitem[{Wu et~al.(2020)Wu, Belinkov, Sajjad, Durrani, Dalvi, and
  Glass}]{wu2020similarity}
John~M Wu, Yonatan Belinkov, Hassan Sajjad, Nadir Durrani, Fahim Dalvi, and
  James Glass. 2020.
\newblock Similarity analysis of contextual word representation models.
\newblock \emph{arXiv preprint arXiv:2005.01172}.

\bibitem[{Yang et~al.(2018)Yang, Liang, and Zhang}]{yang2018design}
Jie Yang, Shuailong Liang, and Yue Zhang. 2018.
\newblock \href {http://aclweb.org/anthology/C18-1327} {Design challenges and
  misconceptions in neural sequence labeling}.
\newblock In \emph{Proceedings of the 27th International Conference on
  Computational Linguistics (COLING)}.

\bibitem[{Yang et~al.(2017)Yang, Zhang, and Dong}]{yang2017neural}
Jie Yang, Yue Zhang, and Fei Dong. 2017.
\newblock Neural word segmentation with rich pretraining.
\newblock In \emph{Proceedings of the 55th Annual Meeting of the Association
  for Computational Linguistics (Volume 1: Long Papers)}, pages 839--849.

\bibitem[{Yang et~al.(2019)Yang, Dai, Yang, Carbonell, Salakhutdinov, and
  Le}]{yang2019xlnet}
Zhilin Yang, Zihang Dai, Yiming Yang, Jaime Carbonell, Ruslan Salakhutdinov,
  and Quoc~V Le. 2019.
\newblock Xlnet: Generalized autoregressive pretraining for language
  understanding.
\newblock \emph{arXiv preprint arXiv:1906.08237}.

\bibitem[{Zampieri et~al.(2018)Zampieri, Malmasi, Nakov, Ali, Shon, Glass,
  Scherrer, Samard{\v{z}}i{\'c}, Ljube{\v{s}}i{\'c}, Tiedemann
  et~al.}]{zampieri2018language}
Marcos Zampieri, Shervin Malmasi, Preslav Nakov, Ahmed Ali, Suwon Shon, James
  Glass, Yves Scherrer, Tanja Samard{\v{z}}i{\'c}, Nikola Ljube{\v{s}}i{\'c},
  J{\"o}rg Tiedemann, et~al. 2018.
\newblock Language identification and morphosyntactic tagging: The second
  vardial evaluation campaign.
\newblock In \emph{Proceedings of the Fifth Workshop on NLP for Similar
  Languages, Varieties and Dialects (VarDial 2018)}, pages 1--17.

\bibitem[{Zhao et~al.(2017)Zhao, Wang, and Li}]{zhao2017deep}
Chuanjun Zhao, Suge Wang, and Deyu Li. 2017.
\newblock Deep transfer learning for social media cross-domain sentiment
  classification.
\newblock In \emph{Chinese National Conference on Social Media Processing},
  pages 232--243. Springer.

\bibitem[{Zhou et~al.(2018{\natexlab{a}})Zhou, Bau, Oliva, and
  Torralba}]{zhou2018interpreting}
Bolei Zhou, David Bau, Aude Oliva, and Antonio Torralba. 2018{\natexlab{a}}.
\newblock Interpreting deep visual representations via network dissection.
\newblock \emph{IEEE Transactions on Pattern Analysis and Machine
  Intelligence}.

\bibitem[{Zhou et~al.(2015)Zhou, Khosla, Lapedriza, Oliva, and
  Torralba}]{zhou2014object}
Bolei Zhou, Aditya Khosla, Agata Lapedriza, Aude Oliva, and Antonio Torralba.
  2015.
\newblock Object detectors emerge in deep scene cnns.
\newblock \emph{ICLR2015}.

\bibitem[{Zhou et~al.(2018{\natexlab{b}})Zhou, Sun, Bau, and
  Torralba}]{zhou2018revisiting}
Bolei Zhou, Yiyou Sun, David Bau, and Antonio Torralba. 2018{\natexlab{b}}.
\newblock Revisiting the importance of individual units in cnns via ablation.
\newblock \emph{arXiv preprint arXiv:1806.02891}.

\bibitem[{Zhu et~al.(2018)Zhu, Li, and De~Melo}]{zhu2018exploring}
Xunjie Zhu, Tingfeng Li, and Gerard De~Melo. 2018.
\newblock Exploring semantic properties of sentence embeddings.
\newblock In \emph{Proceedings of the 56th Annual Meeting of the Association
  for Computational Linguistics (Volume 2: Short Papers)}, pages 632--637.

\bibitem[{Zoph and Knight(2016)}]{zoph2016multi}
Barret Zoph and Kevin Knight. 2016.
\newblock Multi-source neural translation.
\newblock In \emph{Proceedings of NAACL-HLT}, pages 30--34.

\end{thebibliography}
\bibliographystyle{acl_natbib}

\end{document}